\RequirePackage[T1]{fontenc}
\documentclass[letterpaper,journal]{IEEEtran}
\usepackage{amsmath,amssymb,amsfonts}
\usepackage{graphicx}
\usepackage{array,booktabs,multirow,tabularx}
\usepackage{algorithm}
\usepackage{algpseudocode}
\usepackage[caption=false,font=footnotesize]{subfig}
\usepackage{cite}
\usepackage{textcomp}
\usepackage{url}
\usepackage{stfloats}
\usepackage{placeins}
\usepackage[hidelinks,pdfversion=1.7]{hyperref}
\begin{document}
\title{Robust Image Emotion Classification via Multi-Level Affective Captioning}
\author{Zibo Zhou, Zhengjun Zhai, and Huimin Chen%
\thanks{Zibo Zhou and Zhengjun Zhai are with the School of Computer Science, Northwestern Polytechnical University, No.127 Youyi Xilu, Xi'an, Shaanxi 710072, China (e-mail: zibo.zhou@mail.nwpu.edu.cn; zhengjunzhai123@163.com).}%
\thanks{Huimin Chen is with the School of Cybersecurity, Northwestern Polytechnical University, No.127 Youyi Xilu, Xi'an, Shaanxi 710139, China (e-mail: chen\_hm@mail.nwpu.edu.cn).}%
\thanks{Corresponding author: Zhengjun Zhai.}}
\maketitle

\begin{abstract}
Image emotion classification (IEC) has been studied for many years and has recently attracted increasing attention with the rapid development of deep learning. Although recent studies have benefited from knowledge embedded in pre-trained visual models, the performance is still limited by the "affective gap" between low-level visual features and high-level affective semantics. Psychological studies suggest that language can convey fine-grained and emotion-rich information. Inspired by this observation, we attempt to narrow the "affective gap" by introducing affective text. Specifically, we design a multi-level contrastive loss to detect emotional concepts from images and introduce an emotional-attribute-guided large language model approach to generate affective sentences. Then, a pre-trained language model is employed to combine emotional concepts with affective sentences for IEC. In addition, we introduce a contrastive loss with a semantic-similarity-based sampling strategy to address large intra-class differences and limited inter-class differences in affective datasets. Moreover, unlike many previous studies, we explicitly consider images containing embedded text. Extensive experiments on multiple benchmarks show that the proposed method helps narrow the "affective gap" and achieves competitive performance.
\end{abstract}

\begin{IEEEkeywords}
Human cognition, image emotion classification, contrastive loss, large language model, chain-of-thought prompting.
\end{IEEEkeywords}

\section{Introduction}
\label{introduction}
\IEEEPARstart{W}{ith} the development of information technology and the widespread use of social media, images have emerged as a crucial means for people to express their views and gather information. Image Emotion Classification (IEC) seeks to bridge visual content and human emotions, aiming to interpret the affective responses triggered by diverse images. IEC has broad application prospects in public opinion monitoring, user services, and healthcare, making it an emerging and pivotal research topic in computer vision. Compared with other computer vision tasks, IEC is more challenging due to the abstractness and ambiguity of emotions, which result in the "affective gap" between image and emotion.

Grounded in art and psychological theories \cite{siersdorfer2010analyzing,zhao2014exploring}, early IEC approaches relied on hand-crafted features (e.g., color, texture, and shape) to identify the visual factors that influence human emotions. However, the relationship between these explicit features and the evoked emotions remains ambiguous across most scenes. With the advent of deep learning, Convolutional Neural Networks (CNNs) were widely adopted to classify emotions in an end-to-end manner. Yet, as network depth increases, performance improvements often plateau due to the information loss caused by successive pooling operations. Consequently, both traditional feature-based and standard CNN-based methods struggle to fully bridge the "affective gap".

Recently, better performance has been achieved by approaches that utilize intermediate semantics to help establish the relationship between image and emotion. Accordingly, psychological research indicates that people perceive emotions through the Stimulus--Organism--Response (S-O-R) model\cite{mehrabian1974approach}. In traditional and end-to-end methods, the organism (O), which encompasses human perception, experience, and memory is overlooked. Therefore, we believe that intermediate semantics, i.e., the organism in the S-O-R model, are key to understanding image emotion.

However, existing methods incorporating intermediate semantic supervision in IEC still face several limitations. First, some approaches utilize object detection to locate emotional regions \cite{yang2021solver}. However, object detection methods output many overlapping regions, which may induce noises. Second, although some studies employ image captioning as language signals \cite{cen2024masanet}, these methods tend to generate objective descriptions of visual entities rather than capturing emotion-related information. Finally, current methods relying on intermediate semantics supervision frequently use fixed templates combined with emotion labels \cite{deng2022simemotion}, which fail to fully capture and express the rich, nuanced semantic information of an image.

Based on the above observations, we propose Affective Captioning for Image Emotion Classification (ACIEC), which implements IEC via affective intermediate semantics. Specifically, for the strong emotional orientation of adjective-noun pairs (ANPs), which were constructed by Borth et al. \cite{borth2013large}, we first design an ANP detector. Considering the hierarchical relationships among ANP labels, we propose a multi-level contrastive loss to optimize the label space on both noun and ANP levels. Furthermore,  motivated by the achievement of large language model (LLM), we also design emotional attribute chain-of-thought (EA-CoT) prompting to generate affective sentences for images. Subsequently, a pre-trained language model takes both the ANPs and the affective sentences as input to classify the image emotion. Moreover, a contrastive learning strategy based on semantic similarity sampling is introduced to better handle the large intra-class variation and limited inter-class separation of affective images in datasets. Specifically, since some social media images contain embedded text, whose emotions are conveyed primarily through textual rather than visual features, we employ an Optical Character Recognition (OCR) model to extract the text and use LLM to directly infer their emotions. In summary, the main contributions of our work are as follows: \par
\begin{itemize}
\item[1)] We propose ACIEC, which transforms images into multi-level affective textual semantics, including emotional concepts and affective sentences, to help narrow the "affective gap" between visual content and evoked emotions. 
\item[2)] A novel ANP detector is built based on a hierarchical multi-level contrastive objective and a hierarchy-aware sampling strategy, enabling more discriminative affective concept representations. 
\item[3)] We design an emotional-attribute-guided chain-of-thought prompting strategy that guides the LLM to analyze psychologically relevant emotional cues and incorporates a self-consistency strategy to obtain more stable affective sentences.
\item[4)] A semantic-similarity-sampled contrastive learning strategy is designed to alleviate large intra-class variation and limited inter-class separation in affective image datasets.
\item [5)] Extensive experiments on multiple IEC benchmarks demonstrate the improvements achieved by our method, supporting the effectiveness of helping narrow the "affective gap".
\end{itemize}

The remainder of this paper is organized as follows. Section~\ref{related work} reviews the related work. Section~\ref{sec:method} describes our proposed IEC method in detail. In Section~\ref{sec:experiments}, conducted on public affective datasets, various experiments and analysis are presented. Finally, we present our conclusions and discuss future directions in Section~\ref{sec:conclusion}.

\section{Related Work}
\label{related work}
In our study, we investigate the problem of image emotion classification (IEC), which is also related to Chain-of-Thought Reasoning (CoT) and contrastive learning. This section reviews previous studies from these three perspectives. 

\subsection{Image Emotion Classification}
\label{image emotion classification}

Early IEC research mainly depended on low-level visual features. For example, Lu et al. \cite{lu2012shape} explored the emotion-rich content of social media images by statistically analysing shape features such as circles and angles. To investigate the link between art principles and emotions, Zhao et al. \cite{zhao2014exploring} proposed a method for both classification and regression tasks, extracting emotional features based on art principles such as balance, emphasis, harmony, variety, gradation, and movement. Zhan et al. \cite{zhan2019zero} proposed a zero-shot emotion recognition method by constructing an affective structural embedding from adjective--noun pairs and jointly aligning visual and semantic features with adversarial constraints. Zhang et al. \cite{zhang2020object} obtained several objects in each image via object detection, which explored the relationship between object semantic combinations and emotions by Bayesian networks. However, as a high-level semantic signal, emotion is inherently abstract and subjective. Low-level features alone cannot adequately capture all important affective factors, because they lack explicit and interpretable cognitive semantics.

With the development of deep neural networks, many recent studies have explored deep fusion strategies for emotion prediction. Lin et al. \cite{lin2020multi} proposed a multi-source domain adaptation via pixel-level translation to leverage emotion hierarchy. Zhu et al. \cite{zhu2025learning} proposed a prototype-based visual emotion recognition method that jointly learns textual and visual class prototypes to address label ambiguity through prototype-guided alignment and distance-based label smoothing. Furthermore, Zhou et al. \cite{zhou2025object} proposed an OSAFN which extracts object-level concepts and generates scene descriptions, then aligns semantics with visual regions through dedicated attention modules and integrates the two semantic streams using an adaptive gating fusion mechanism.

Building on these observations, we develop a novel IEC framework that incorporates multi-level language information motivated by psychological findings. Specifically, a multi-level contrastive loss is designed to detect emotional concepts, while an emotional attribute chain-of-thought approach generates affective sentences. Finally, a pre-trained language model is used for the final classification.

\subsection{Chain-of-Thought Reasoning}
\label{CoT}

Chain-of-Thought Reasoning (CoT) guides a large language model (LLM) to decompose complex problems into manageable subtasks through step-by-step reasoning. Existing methods fall into two categories: few-shot and zero-shot. Few-shot CoT relies on manually annotated templates. For instance, Wei et al. \cite{wei2022chain} provided a sequence of reasoning steps in the prompt, enabling significant gains in arithmetic, common-sense, and symbolic reasoning without any fine-tuning process. In contrast, for zero-shot CoT prompting, task-agnostic triggers are designed to prompt large language models to produce step-by-step reasoning processes. Wang et al. \cite{wang2025multimodal} proposed PS-CoT, which guides an LLM to decompose the task into manageable sub-tasks and carry them out step by step, and Kojima et al. \cite{kojima2022large} demonstrated that even simple instructions such as ``Let us consider step by step'' can effectively elicit reasoning.

Recently, CoT has also been applied to emotion-related tasks. Wu et al. \cite{wu2024enhancing} proposed a deconstructed reasoning framework that enables an LLM to extract emotion--cause pairs through stepwise inference. Besides, Li et al. \cite{li2024enhancing} enhanced the emotional generation capacity of LLM by incorporating an emotional-CoT, allowing models to more effectively embed affective elements in their responses. Our method extends these insights by integrating a psychological theory to construct emotional attributes, improving the LLM's understanding of emotional content in visual scenes.

\subsection{Contrastive Learning}
\label{cl}

Contrastive learning aims to learn discriminative representations by minimizing the distance between positive pairs in the embedding space while maximizing the distance between negative pairs. Seminal frameworks such as MoCo \cite{he2020momentum} and SimCLR \cite{chen2020simple} have driven significant progress in this area: MoCo leverages a momentum encoder to maintain a large dictionary of negatives, whereas SimCLR emphasizes the construction of diverse positive pairs. Most existing methods rely heavily on data augmentation pipelines---such as random cropping, rotation, colour jittering, and Sobel filtering---to generate these positive pairs. Such operations compel the model to capture invariant semantics across diverse augmented views, thereby enhancing feature extraction. To further enrich positive samples, Fan et al. \cite{fan2021unsupervised} introduced a dual-threshold scheme. Furthermore, when label information is available, pairs can be constructed directly via class semantics: samples sharing a label are treated as positive pairs, while those from different classes form negative pairs. Building on this, Zhong et al. \cite{zhong2021neighborhood} proposed generating novel samples by integrating labelled and unlabelled instances within the feature embedding space.

\begin{figure*}[!t]
  \centering
  \includegraphics[pagebox=mediabox,trim=71.87bp 16.37bp 102.80bp 35.54bp,clip,width=\textwidth]{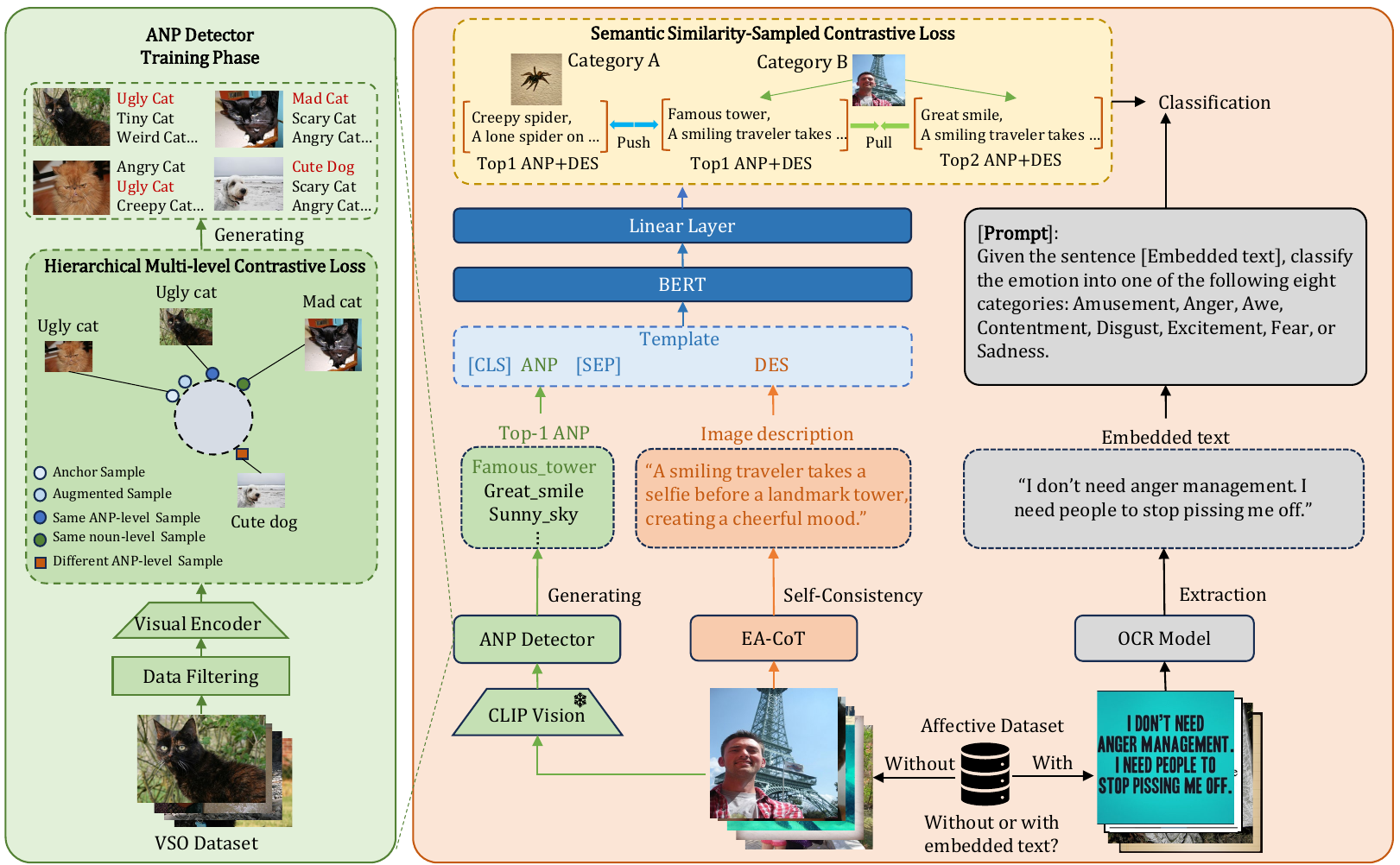}
  \caption{The pipeline of the proposed ACIEC.}
  \label{framework}
   
\end{figure*}

\section{Proposed Method}
\label{sec:method}

In this section, we elaborate on the details of ACIEC. In contrast to existing approaches, we reformulate IEC as an emotion classification problem based on affective intermediate semantics, where visual content is first converted into emotion-related textual representations before final prediction. The pipeline of ACIEC is demonstrated in Fig.~\ref{framework}. Given an input image, embedded texts are first detected through an optical character recognition (OCR) model. For images with embedded texts, a zero-shot prompt is designed to predict emotion directly based on the extracted texts. For other images without embedded texts, corresponding ANP and affective sentences are generated by the ANP detector and emotional attribute chain-of-thought (EA-CoT), respectively. Finally, another pre-trained language model is utilized to combine the ANP and affective sentence for classification. A novel contrastive loss based on semantic similarity sampling is also introduced to address the large intra-class differences and small inter-class differences in affective datasets. Further elaboration of these parts is in the subsequent subsections.

\subsection{ANP Detector}
\label{ad}

The existing ANP detector, DeepSentiBank \cite{chen2014deepsentibank}, suffers from two main limitations: (1) the VSO dataset \cite{borth2013large} which is used for training the detector, contains a large number of ANP-irrelevant images, which hinders the performance; and (2) it relies on the outdated Caffe framework \cite{jia2014caffe}, limiting feature representation compared with modern deep models. To address these issues, we propose a novel ANP detector based on a multi-level contrastive loss. The details are as follows. 

\begin{figure}[!t]
\centering
\includegraphics[pagebox=mediabox,trim=229.73bp 301.81bp 368.40bp 119.68bp,clip,width=\columnwidth]{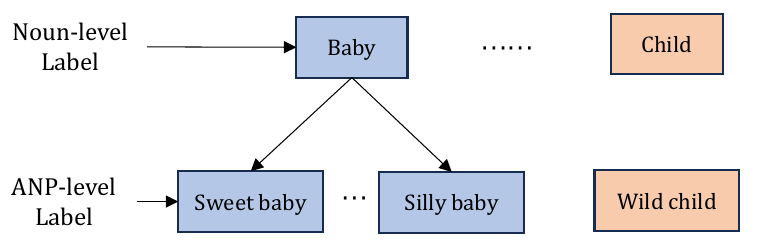} 
  \caption{Example of the hierarchical structure of the VSO dataset.}
  \label{example}
   
\end{figure}

\subsubsection{VSO Data Filtering}
\label{df}

The VSO dataset contains approximately 1,000,000 images and 3,000 emotional concepts. However, the images collected for each concept were not manually screened; the irrelevant images restrict the performance of DeepSentiBank. Since semantic consistency between image and concept is the key to determining the relevance of an image with the corresponding concept, we used a pre-trained visual-language model to estimate the correlation between image and emotional concepts. Specifically, the visual encoder of CLIP \cite{radford2021learning} takes an image $V$ as input and outputs the embedded visual feature $F_v$. Since the content of the image is primarily determined by the noun, we utilize the same sentence template "A photo of a []" in CLIP, and the nouns are placed into the placeholder "[]". The template is then put into the CLIP text encoder, producing an output feature $F_t$. The similarity between the image and the text is calculated by cosine similarity, which is denoted by $S_{vt}$:

\begin{equation}
    S_{vt} = \frac{F_v\cdot F_t}{||F_v|| \cdot ||F_t||}
\end{equation}

The threshold was empirically selected to remove obviously irrelevant samples while preserving sufficient training dat. If $S_{vt}$ is less than 0.95, the image $V$ is considered to be irrelevant to its corresponding ANP and removed from the VSO dataset.

\subsubsection{Hierarchical Multi-Level Contrastive Learning}
The goal of contrastive learning is to minimize the distance between an anchor and its positive samples in the embedding space while maximizing its distance from negative samples. Specifically, given a mini-batch $B = \{x_i \mid i \in I = \{1, \ldots, BS\}\}$, where $BS$ represents the size of the batch, the loss is defined as follows:

\begin{equation}
\begin{aligned}
L &= \sum_{i \in I} \frac{-1}{|\Gamma(i)|} \sum_{\mu \in \Gamma(i)} \\
&\quad {}\times \log \frac{\exp \!\left( \mathrm{score}\!\left(v_i,\, v_\mu\right) / \tau \right)} {\sum_{\eta \in \Lambda \setminus \{i\}} \exp \!\left( \mathrm{score}\!\left(v_i,\, v_\eta \right)/ \tau \right)}
\end{aligned}
\end{equation}

where $\tau$ represents the temperature parameter, $\Gamma(i)$ represents the indices of all positive samples in the batch except for $i$, $\Lambda$ represents all images in the batch,  and $\eta \in \Lambda \setminus \{i\}$ represents all images in the batch except the anchor image.

Since an emotional concept is composed of a noun and an adjective, it is obvious that for the emotional concepts with the same noun label, the corresponding images exhibit high similarity because the noun reflects the physical presence of objects or scenes in the image. Hence, we argue that ANP labels in the VSO dataset have a hierarchical structure. For example, in Fig.~\ref{example}, leaf nodes correspond to emotional concepts with the same noun (sweet baby, silly baby, wild child), and the root corresponds to the respective noun (baby, child).

Although the formulation in Eq.~(2) effectively aligns multiple positive pairs, it operates exclusively on single-level annotations, thereby neglecting the hierarchical relationships among emotional concepts. In order to use the hierarchical structure of labels to better distinguish the sample space, we propose a multi-level contrastive loss function:

\begin{equation}
    L_{\mathrm{con}} =\frac{1}{2}\Bigl(L_{\mathrm{con}}^{\mathrm{noun}}+L_{\mathrm{con}}^{\mathrm{ANP}}\Bigr)
\end{equation}

where $L_{\mathrm{con}}^{\mathrm{noun}}$ represents noun-level contrastive loss, which can be defined as:

\begin{equation}
\begin{aligned}
L_{\mathrm{con}}^{\mathrm{noun}} &= \sum_{i \in I} \frac{-1}{\left|\Gamma^{\mathrm{noun}}(i)\right|} \sum_{\mu \in \Gamma^{\mathrm{noun}}(i)} \\
&\quad {}\times \log \frac{\exp \!\left( \mathrm{score}\!\left(v_i,\, v_{\mu}^{\mathrm{noun}} \right) / \tau \right)} {\sum_{\eta \in \Lambda \setminus \{i\}} \exp \!\left( \mathrm{score}\!\left(v_i,\, v_{\eta}^{\mathrm{noun}} \right) / \tau \right)}
\end{aligned}
\end{equation}

Here, $v_{\mu}^{\mathrm{noun}}$ represents the positive samples with the same noun-level label as the anchor sample and $\Gamma^{\mathrm{noun}}(i)$ is defined as the set of positive samples with the same noun-level label as the anchor sample. The score is defined as the cosine similarity of representation vectors.  

Besides, the ANP-level contrastive loss is defined as:

\begin{equation}
\begin{aligned}
L_{\mathrm{con}}^{\mathrm{ANP}} &= \sum_{i \in I} \frac{-1}{|\Gamma^{\mathrm{ANP}}(i)|} \sum_{\mu \in \Gamma^{\mathrm{ANP}}(i)} \\
&\quad {}\times \log \frac{\exp\!\left( \mathrm{score}\!\left(v_i,\, v_{\mu}^{\mathrm{ANP}} \right) / \tau \right)} {\sum_{\eta \in \Lambda \setminus \{i\}} \exp\!\left( \mathrm{score}\!\left(v_i,\, v_{\eta}^{\mathrm{ANP}} \right)/ \tau \right)}
\end{aligned}
\end{equation}

Here, $v_{\mu}^{\mathrm{ANP}}$ represents the positive samples with the same ANP-level label as the anchor sample and $\Gamma^{\mathrm{ANP}}(i)$ is defined as the set of positive samples with the same ANP-level label as the anchor sample.  

A fully connected layer that outputs the probability distribution of the ANP category is added after the visual model. The loss between the output and the ANP label is calculated by the cross-entropy loss function. Finally, the loss function used to optimize the ANP detection model is:

\begin{equation}
     L_{\mathrm{ANP}}^{\mathrm{final}} = L_{\mathrm{CE}}^{\mathrm{ANP}} + L_{\mathrm{con}}
\end{equation}

\subsubsection{Hierarchy-Aware Batch Sampling}
\label{hbss}

To encourage diverse multi-level representations within each batch, our sampling strategy pairs every anchor image with positive samples sharing common ancestors in the hierarchy. First, we randomly select an anchor image $X_a$ with its complete label hierarchy $\mathcal{H}=\{L_{\text{noun}}, L_{\text{ANP}}\}$, where $L_{\text{noun}}$ denotes the noun label, and $L_{\text{ANP}}$ denotes the ANP label. Next, we sample another image $X_p$ from the same subcategory $L_{\text{ANP}}$ to form a positive pair $(X_a,X_p)$. Then, we select an image $X_n$ from different subcategories within $L_{\text{noun}}$, creating a hierarchical contrastive relationship where $X_n$ serves as (1) a negative sample for $L_{\text{ANP}}$--level discrimination and (2) a positive sample for $L_{\text{noun}}$--level representation learning. Finally, we repeat these processes until either all subcategories within the major category have been sampled or the desired batch size of images has been collected. If all subcategories are sampled but the batch size has not been reached, we select a new anchor image and restart the random sampling process. If the batch size is reached before all subcategories are sampled, we stop the sampling process.To prevent redundant sampling, each image is strictly sampled only once per epoch. The overall procedure is detailed in Algorithm~\ref{algorithm1}.

\begin{algorithm}[!t]
\footnotesize
\caption{Hierarchical Contrastive Sampling Strategy}
\label{algorithm1}
\begin{algorithmic}[1]
\State \textbf{Input:} Batch size $B$, dataset with labels $L_{\text{noun}}, L_{\text{ANP}}$
\State $\mathcal{B}\leftarrow \emptyset$; mark all images as unvisited
\While{$|\mathcal{B}|<B$}
  \State Randomly select anchor $X_a$ with $(L_{\text{noun}}(X_a),L_{\text{ANP}}(X_a))$
  \State Select $X_p$ with $L_{\text{ANP}}(X_p)=L_{\text{ANP}}(X_a)$
  \State Select $X_n$ with $L_{\text{noun}}(X_n)=L_{\text{noun}}(X_a)$ and $L_{\text{ANP}}(X_n)\neq L_{\text{ANP}}(X_a)$
  \State $\mathcal{B}\leftarrow \mathcal{B}\cup \{(X_a,X_p,X_n)\}$; mark $X_a,X_p,X_n$ visited
  \If{all sub-categories in $L_{\text{noun}}(X_a)$ are visited and $|\mathcal{B}|<B$}
     \State \textbf{continue}
  \EndIf
\EndWhile
\State \textbf{Constraint:} Each image is sampled at most once per epoch
\end{algorithmic}
\end{algorithm}

\subsection{Emotional Attribute Chain-of-Thought}
\label{EA-CoT}

Due to the "affective gap" between visual features and emotions, most image captioning methods generate plain image captions while missing emotion-related information, limiting the effectiveness of IEC. To address this issue, we introduce an Emotional Attribute Chain-of-Thought (EA-CoT), which uses several psychological theories to guide the LLM through an explicit affective reasoning process.

\subsubsection{Emotional-Attribute Prompt Construction}
\label{pc}

First, we defined four emotional attributes based on psychological studies; the details are as follows:
\begin{itemize}
\item[1)]
\textbf{Scene}: Brosch et al. \cite{brosch2010perception} suggested that scenes can serve as direct emotional stimuli in images. The scene plays a crucial role in evoking emotions, particularly when images lack salient objects such as people, animals, or distinct items. Consequently, a comprehensive caption of the scene can effectively elicit a wide range of emotional responses from the viewer.
\item[2)]
\textbf{Object}: Steward et al. \cite{steward2025interactions} indicated that objects carry stable affective valence which have influence on perception, attention, and memory. Besides, the perception of emotions is integrative, incorporating information from both object and surroundings.
\item[3)]
\textbf{Facial expression}: Ekman \cite{ekman1993facial} demonstrated that facial expressions, served as an important source of affective information, offer explicit visual cues for recognizing emotional states.
\item[4)]
\textbf{Human action}: Gelder et al. \cite{de2015perception} found that viewer can infer emotion from body expressions, such as posture, movement, and action tendencies. We consider events as part of human action in our study. 
\end{itemize}

\begin{figure}[!t] 
  \centering
  \includegraphics[pagebox=mediabox,trim=272.40bp 289.03bp 307.60bp 74.95bp,clip,width=\columnwidth]{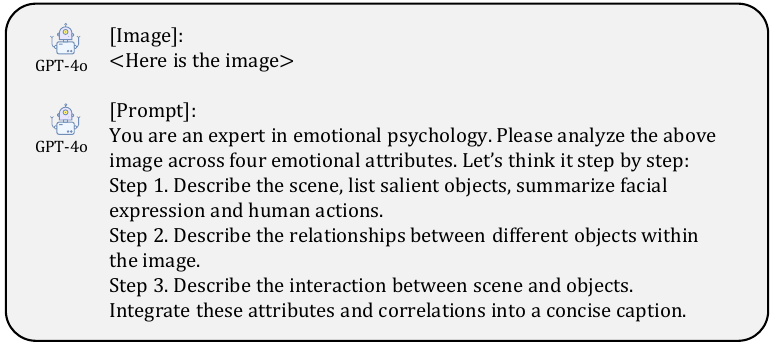} 
  \caption{A template of an EA-CoT prompt.}
  \label{prompt}
\end{figure}

Psychologists indicated that emotions arise from the perception of meaningful objects rather than any single object \cite{frijda2009emotion}. Similarly, neuroscientist emphasizes that visual objects are processed within rich surroundings and co-occur with related objects, which is a key cognitive operation in the human brain \cite{bar2004visual}. Building on these insights, we design EA-CoT as a three-step reasoning scheme that makes these interactions before producing an affective sentence. The overall EA-CoT is illustrated in Fig.~\ref{prompt}.

Compared with our previous work \cite{zhou2025object}, EA-CoT explicitly decomposes affective evidence into four psychologically grounded attributes, and further surfaces "object-object" relations and "scene-object" interactions before generating the affective sentence.

\subsubsection{Self-Consistency Aggregation}
To further improve the reliability of EA-CoT, we adopt a self-consistency strategy, following the idea proposed in \cite{wang2022self}. The goal is to reduce the instability that may arise from a single reasoning process and improve the accuracy and reliability of the result. The self-consistency approach follows two steps:
\begin{itemize}
\item[1)] 
\textbf{Generating multiple reasoning chains}: Given an image, the LLM is prompted to generate multiple reasoning chains. Each chain may emphasize different object relations or contextual cues, thereby providing diverse affective sentences of the same image.
\item[2)] 
\textbf{Evaluation and aggregation}: After generating multiple reasoning chains, the model evaluates the affective sentences derived from these chains and selects the most consistent result. Following the self-consistency principle, the affective sentence that appears most frequently across the $K$ reasoning paths is chosen as the final output. In our implementation, $K$ is set to 3 to balance robustness and computational cost. 
\end{itemize}

By integrating multiple reasoning paths, the self-consistency approach mitigates the errors and biases that may arise from a single CoT, thereby enhancing the robustness and accuracy of the overall reasoning process. 

\begin{figure*}[!t]
  \centering
  \includegraphics[pagebox=mediabox,trim=204.13bp 116.49bp 181.73bp 76.01bp,clip,width=\textwidth]{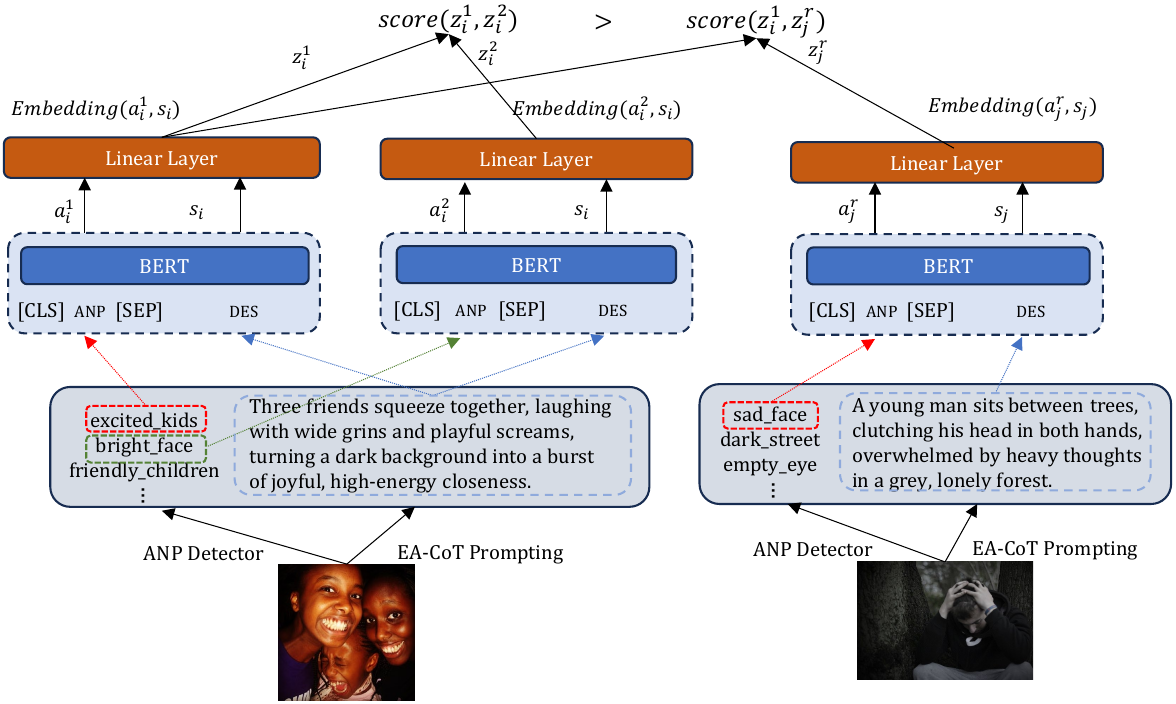}
  \caption{The sampling procedure of the proposed loss function.}
  \label{loss}
  
\end{figure*}

\subsection{Emotion Prediction with Multi-Level Affective Semantics}

Existing methods generally regard visual features as the primary source of information for image emotion classification. However, images used in IEC tasks are commonly collected from social media and therefore encompass a wide variety of content, including embedded text. When treated merely as part of the visual appearance, such text may interfere with visual feature extraction and hinder emotion recognition. Nevertheless, once explicitly recognized, the embedded text can provide semantic and emotional cues that are often consistent with the visual content. We therefore process text-embedded and text-free images separately.

Specifically, we employ an OCR model \cite{cui2025paddleocr} to identify images containing embedded text and extract their textual content. Under the same OCR detection criterion, text-embedded images account for 15.74\% of FI, 20.81\% of EmotionROI, and 24.51\% of Twitter I. These proportions also represent the percentages of images processed through the OCR-based branch in the corresponding datasets. Because text-embedded images constitute only a subset of each dataset and their extracted texts vary considerably in length and content, directly training a supervised BERT-based text classifier may lead to unstable learning and insufficient utilization of the textual information. To address this issue, we employ a zero-shot prompting strategy to directly infer the emotions conveyed by the extracted text. The detailed prompt is presented in Fig.~\ref{framework}.

For images without embedded text, ANPs are highly abstract and exhibit clear emotional tendencies, whereas affective sentences can capture finer-grained image semantics. To leverage the complementarity of these two types of semantics, we employ BERT \cite{devlin2019bert} for IEC. Based on the generated $ANP$ and $Des$, the sentence template is constructed as T(ANP, Des) = \{ [CLS]\ ANP\ [SEP]\ Des \} , where [CLS] and [SEP] are two special tokens for classification and separation, respectively. $ANP$ denotes emotional concepts and $Des$ denotes affective sentences. The hidden state corresponding to the [CLS] token $h_{cls}$ is treated as the final representation that is fed into a softmax layer to compute the probability distribution. 

\subsection{Semantic Similarity-Sampled Contrastive Loss}
\label{lf}

Considering the problem of large intra-class differences and small inter-class differences in affective datasets, we propose contrastive learning based on a semantic similarity sampling strategy in which sampling is conducted via intermediate semantics. Fig.~\ref{loss} illustrates the overall sampling process of our loss, for each image \(x_i\) we obtain a ranked set of \(K\) ANPs \(A_i=\{a_i^{k}\}_{k=1}^{K}\) with probabilities \(\{\pi_i^{(k)}\}\) and an affective sentence \(s_i\). We form the anchor pair \((a_i^{1}, s_i)\) using the top-1 ANP and the positive pair \((a_i^{2}, s_i)\) using the top-2 ANP from the same image. Negative samples are constructed by randomly sampling ANP-sentence pairs \((a_{r}^{\,j}, s_j)\) from images \(j\neq i\) belonging to other categories (\(a_{r}^{\,j}\in A_j\)). These pairs are then used in a contrastive loss. We define the embedding features of each sample as:

\[
\begin{aligned}
\mathbf{z}_i^{1} &= \text{Emb}(a_i^{1},\, s_i),\\
\mathbf{z}_i^{2} &= \text{Emb}(a_i^{2},\, s_i),\\
\mathbf{z}_j^{r} &= \text{Emb}(a_j^{r},\, s_j).
\end{aligned}
\]

Then, the contrastive loss for each anchor--positive pair is formulated as:

\begin{equation}
\label{eq:infonce_pair}
\begin{aligned}
\mathcal{L}_{\text{con}}
&= -\log
\frac{\exp\!\left(\mathrm{score}\!\left(\mathbf{z}_i^{1},\,\mathbf{z}_i^{2}\right)/\tau\right)}
{\begin{aligned}
&\exp\!\left(\mathrm{score}\!\left(\mathbf{z}_i^{1},\,\mathbf{z}_i^{2}\right)/\tau\right)\\
&{}+\sum_{j\neq i}\sum_{r}\exp\!\left(\mathrm{score}\!\left(\mathbf{z}_i^{1},\,\mathbf{z}_j^{r}\right)/\tau\right)
\end{aligned}}
\end{aligned}
\end{equation}

where $\tau$ is a temperature parameter controlling the sharpness of the distribution. This formulation maximizes the similarity between the anchor $(a_i^1, s_i)$ and its positive $(a_i^2, s_i)$ while minimizing similarities to random negatives.

Subsequently, when given a batch of images, the total loss can be defined as the sum of classification loss and contrastive loss. The two loss functions are combined with a weighting coefficient $\lambda$, which is illustrated as follows:

\begin{equation}
    \mathcal{L} = \mathcal{L}_{\text{CE}} + \lambda \mathcal{L}_{\text{con}}
\end{equation}

where $\lambda$ balances the contribution of classification and contrastive learning.

\section{Experiments and Evaluation}
\label{sec:experiments}

\subsection{Datasets and Evaluation Protocol}
Our experiments are conducted on four public affective datasets collected from social media, namely Twitter I \cite{you2015robust}, Twitter II \cite{borth2013large}, EmotionROI \cite{peng2015mixed}, and FI \cite{you2016building}. These datasets contain abundant emotionally expressive content and exhibit a strong correlation between visual cues and affective semantics, thereby providing a solid foundation for training and evaluating models designed for image emotion classification (IEC). The details are as follows.
\begin{itemize}
\item[1)] 
\textbf{Twitter I} contains 1,269 images annotated with positive or negative emotions following a binary emotion model. Each image was annotated by five Amazon Mechanical Turk (AMT) participants. The final label for an image was determined by majority voting, where the emotion category receiving at least three votes was assigned as the ground-truth annotation.
\item[2)]  
\textbf{Twitter II} was annotated using the same AMT protocol, but through three distinct rounds of labeling: text-based, image-based, and text--image-based. In each round, three AMT participants were invited to label the images, and an image was considered agreed upon if at least two annotators selected the same label. 
\item[3)]
\textbf{EmotionROI} was retrieved from Flickr using six basic emotion keywords (surprise, joy, disgust, fear, sadness, and anger). To ensure that the selected images evoke emotions through low-level visual features, images containing obvious facial expressions or explicit emotional text were removed from the dataset. Finally, 1,980 images were obtained. 
\item[4)]
\textbf{Flickr \& Instagram (FI)} was collected from Flickr and Instagram based on the Mikels emotion model \cite{mikels2005emotional}. The researchers first collected three million weakly labeled emotional images from Flickr and Instagram, then filtered them based on label consistency and class balance. Images in FI were annotated by five AMT participants. Each participant selected one emotion from eight categories, and a label was assigned to an image only if at least three workers agreed on the same category.
\end{itemize}

\subsection{Implementation Details}
The images are resized to $224 \times 224$ pixels. To ensure that the main information of the images is preserved during augmentation, for images with an aspect ratio greater than 1, 10\% of the left or right side is randomly cropped, while for images with an aspect ratio less than 1, 10\% of the top or bottom side is randomly cropped. The dataset is randomly split into 80\% for training, 5\% for validation, and 15\% for testing. Model optimization is performed using stochastic gradient descent (SGD) with a batch size of 64 and a learning rate of 0.001. 

For all LLM-based caption generation, we used GPT-4o-image. The sampling temperature was set to 0.2, with top-p = 1.0 and maximum output tokens = 120. We used one fixed EA-CoT prompt throughout the experiments. No prompt tuning was performed on the test set. For images without embedded text, each image required 3 LLM calls for EA-CoT generation and 1 additional call for aggregation. For images with embedded text, OCR was first applied and one zero-shot LLM call was used for emotion prediction based on the extracted text. Overall, the total number of LLM calls was 27,160. The total LLM inference time was approximately 11.2 hours, and the estimated API cost was 130\$. The trainable components were optimized on two NVIDIA RTX 4090 GPUs, requiring approximately 6.8 hours for ANP detector training and 0.5 hours for RoBERTa-based emotion classifier training \cite{liu2019roberta}.

\subsection{Comparison with Existing Methods}
\subsubsection{Comparison with Competitive IEC Baselines}
To validate the effectiveness of our proposed framework, we compare it with a series of state-of-the-art IEC methods. The experiments are carried out in four benchmark settings, including FI (8-class and 2-class), EmotionROI (6-class and 2-class), Twitter I, and Twitter II. Detailed information on these methods is introduced as follows:

\begin{itemize}
\item[1)] 
\textbf{WSCNet} \cite{she2019wscnet} integrated a detection branch that generates a sentiment map using a cross-spatial pooling strategy, and a classification branch that couples this map with deep features to leverage both holistic and localized information for emotion classification.
\item[2)] 
\textbf{SOLVER} \cite{yang2021solver} constructed an emotional graph to model the relationships between different objects and the correlation between objects and the scene for emotion classification.
\item[3)] 
\textbf{MSRCA} \cite{zhang2022image} detected object features within an image and leveraged a self-attention mechanism to capture the semantic relationships among them, producing a unified representation of object-level semantics for emotion understanding.
\item[4)] 
\textbf{OEAN} \cite{zhang2024object} combined object-guided visual attention and semantic modeling of object-emotion mappings, and fused both modalities via a BiGRU to improve emotion classification.
\item[5)] 
\textbf{DeepSentiBank} \cite{chen2014deepsentibank} developed an emotional concept detector for identifying 1,200 adjective-noun pairs (ANPs) in images, and these detected ANPs were used as mid-level representations for IEC.
\item[6)] 
\textbf{UGRIE} \cite{shi2023one} extracted visual features and constructed sentence templates containing affective words. The two modalities were then fused and encoded using the encoder of the BART model. Finally, the output tokens from the BART decoder were used to predict the emotions.
\item[7)] 
\textbf{CMANet} \cite{yang2024concept} utilized emotional concepts to extract visual and semantic features, respectively. Two fusion strategies were applied to achieve complementation between visual and semantic features.
\item[8)] 
\textbf{MASANet} \cite{cen2024masanet} focused on integrating specific unimodal tasks into a unified multi-modal emotion analysis framework by incorporating prior knowledge from textual domains.
\item[9)] 
\textbf{DMSP-KA} \cite{chen2025dynamic} proposed a dynamic multi-scenario prompt learning framework augmented with external knowledge, which adaptively optimizes prompts across diverse contexts to improve the accuracy and robustness of image emotion analysis.
\item[10)] 
\textbf{SimEmotion} \cite{deng2022simemotion} was a language-supervised model that effectively leveraged the rich semantics of the image and text of CLIP, combining visual and textual features to enhance emotional perception with language prompts.
\item[11)] 
\textbf{CoOp} \cite{zhou2022learning} modeled a prompt's context words with continuous learnable vectors to automate prompt engineering and adapted large pre-trained vision-language models for downstream image recognition tasks.
\item[12)] 
\textbf{PT-DPC} \cite{deng2024learning} utilized affective words and visual features to generate dynamic template prefixes for constructing sentence templates. Then the CLIP model was utilized to match visual features with the encoded sentence templates to train the model.
\end{itemize}

\begin{table*}[!t]
\centering
\small
\setlength{\tabcolsep}{4pt}
\renewcommand{\arraystretch}{1.10}
\renewcommand{\tabularxcolumn}[1]{m{#1}}

\caption{Comparison with state-of-the-art IEC methods on different IEC datasets.}
\label{comparison}
\begin{tabular*}{\linewidth}{@{\extracolsep{\fill}}lcccccc@{}}
\toprule
\textbf{Method} & \multicolumn{2}{c}{\textbf{FI}} & \multicolumn{2}{c}{\textbf{EmotionROI}} & \textbf{Twitter I} & \textbf{Twitter II} \\
\cmidrule(lr){2-3} \cmidrule(lr){4-5}
\textbf{Number of classes} & \textbf{8} & \textbf{2} & \textbf{6} & \textbf{2} & \textbf{2} & \textbf{2} \\
\midrule
WSCNet          & 70.07          & --             & 58.25            & --               & 84.25              & 81.35 \\
SOLVER          & 72.33          & --             & 62.12            & --               & 85.43              & 83.19 \\
MSRCA           & 72.60          & 87.40          & 55.60            & 83.00            & --                 & --    \\
OEAN            & 73.44          & --             & 64.14            & --               & 85.59              & --    \\
DeepSentiBank   & 51.54          & 61.54          & 42.53            & 70.11            & 71.25              & 70.23 \\
UGRIE           & 76.48          & 92.48          & 68.18            & 87.04            & 93.70              & 86.47 \\
CMANet          & 73.21          & 89.86          & 62.29            & 84.18            & 85.95              & --    \\
MASANet         & 79.16          & 95.07          & 73.23            & 91.75            & 92.16              & 88.70 \\
DMSP-KA         & 80.68          & --             & 73.74            & --               & 92.13              & 88.72 \\
SimEmotion      & 80.53          & 94.66          & 68.86            & 90.40            & 90.55              & 83.46 \\
CoOp            & 78.66          & 91.93          & 68.48            & 89.74            & 89.05              & 83.78 \\
PT-DPC          & 78.07          & 93.89          & 69.70            & 88.55            & 90.94              & 82.50 \\

\midrule
\textbf{ACIEC}   & \textbf{81.98} & \textbf{95.68} & \textbf{74.66}   & \textbf{92.14}   & \textbf{94.54}     & \textbf{91.86} \\
\bottomrule
\end{tabular*}
\par
\raggedright
\footnotesize \textit{Note:} Results are reported as classification accuracy (\%) on four affective datasets.
\end{table*}

\begin{table*}[!t]
\centering
\small
\setlength{\tabcolsep}{4pt}
\renewcommand{\arraystretch}{1.10}
\renewcommand{\tabularxcolumn}[1]{m{#1}}

\caption{Class-wise evaluation metrics on different IEC datasets.}
\label{tab:combined_metrics}
\begin{tabular*}{\linewidth}{@{\extracolsep{\fill}}llcccc@{}}
\toprule
\textbf{Dataset} & \textbf{Category} & \textbf{Precision} & \textbf{Recall} & \textbf{F1} & \textbf{Specificity} \\
\midrule
\multirow{9}{*}{FI} 
 & amusement   & 90.86 & 88.07 & 89.44 & 97.52 \\
 & anger       & 78.14 & 72.09 & 74.99 & 98.86 \\
 & awe         & 79.25 & 85.29 & 82.16 & 96.36 \\
 & contentment & 81.95 & 81.48 & 81.71 & 94.28 \\
 & disgust     & 81.56 & 87.28 & 84.32 & 98.45 \\
 & excitement  & 76.55 & 80.12 & 78.29 & 96.72 \\
 & fear        & 80.12 & 53.01 & 63.81 & 99.42 \\
 & sadness     & 77.74 & 81.16 & 79.41 & 97.03 \\
 \cmidrule{2-6}
 & \textbf{Average} & \textbf{80.77} & \textbf{78.56} & \textbf{79.27} & \textbf{97.33} \\
\midrule
\multirow{7}{*}{EmotionROI} 
 & anger    & 80.60 & 71.60 & 75.83 & 95.89 \\
 & disgust  & 82.60 & 81.20 & 81.89 & 97.41 \\
 & fear     & 75.60 & 63.60 & 69.08 & 95.58 \\
 & joy      & 78.50 & 77.50 & 78.00 & 94.50 \\
 & sadness  & 77.30 & 80.70 & 78.96 & 95.96 \\
 & surprise & 57.60 & 76.20 & 65.61 & 90.31 \\
 \cmidrule{2-6}
 & \textbf{Average} & \textbf{75.37} & \textbf{75.13} & \textbf{74.90} & \textbf{94.94} \\
\midrule
\multirow{3}{*}{Twitter I} 
 & Positive & 93.74 & 97.30 & 95.49 & 90.50 \\
 & Negative & 95.82 & 90.50 & 93.08 & 97.30 \\
 \cmidrule{2-6}
 & \textbf{Average} & \textbf{94.78} & \textbf{93.90} & \textbf{94.29} & \textbf{93.90} \\
\midrule
\multirow{3}{*}{Twitter II} 
 & Positive & 97.92 & 91.50 & 94.60 & 93.13 \\
 & Negative & 75.61 & 93.13 & 83.46 & 91.50 \\
 \cmidrule{2-6}
 & \textbf{Average} & \textbf{86.77} & \textbf{92.32} & \textbf{89.03} & \textbf{92.32} \\
\bottomrule
\end{tabular*}
\end{table*}

Table~\ref{comparison} reports the comparison results on four affective datasets, where ACIEC achieves the highest accuracy across all evaluation settings. Specifically, traditional object-based methods, including WSCNet, SOLVER, MSRCA, and OEAN, aim to improve IEC by detecting salient regions or objects. However, affective cues are not always explicitly associated with objects. Moreover, detecting a large number of objects may produce overlapping regions, and inappropriate detection thresholds can introduce noise. These limitations make it difficult for object-based methods to reliably capture high-level emotional semantics. 

Compared with multimodal fusion methods, UGRIE improves IEC by incorporating image captions. Nevertheless, it mainly concatenates visual and textual features, which may not sufficiently exploit the complementary relationships between modalities. Similarly, CMANet introduces emotional concepts to extract visual and semantic features, but its fusion strategy remains limited in capturing deeper affective associations. CoOp and PT-DPC further use prompt or emotional templates based on vision-language models. However, these methods mainly align images with predefined textual prompts or templates, and thus still face difficulties in bridging the "affective gap" between low-level visual features and high-level human emotions.

Recent methods such as SimEmotion, MASANet, and DMSP-KA benefit from language supervision and achieve competitive performance on several datasets. However, ACIEC achieves the highest accuracy among the compared methods. For example, compared with MASANet, ACIEC improves the accuracy by 2.82\% on the 8-class setting of FI, 1.43\% on the 6-class setting of EmotionROI, and 3.16\% on Twitter II. These results indicate that simply introducing captions, prompts, or external knowledge remains insufficient for comprehensive affective understanding.

Furthermore, we also observe that our model shows less improvement on the EmotionROI dataset compared to FI. The likely reason is that the RoBERTa-based language encoder is parameter-heavy and benefits from the large size of FI, whereas EmotionROI (1,980 images) is too small to adequately optimize it, leading to under-training and limited improvement. 

Table~\ref{tab:combined_metrics} reports the detailed evaluation results on the four benchmarks. On FI, the F1-score ranges from 89.44\% for amusement to 63.81\% for fear, indicating that some emotion categories are more difficult to recognize than others. Categories such as amusement, disgust, awe, and contentment obtain relatively high F1-scores, whereas fear has a much lower recall of 53.01\%, suggesting that many fear samples are misclassified as other emotions. In addition, amusement achieves both high precision and high recall, whereas fear and anger show relatively high precision but lower recall. In contrast, awe, disgust, excitement, and sadness have higher recall than precision, indicating that these categories are more frequently retrieved but may also include more false positives.

On EmotionROI, surprise category has a relatively high recall of 76.20\% but a low precision of 57.60\%, indicating that many samples from other categories are incorrectly predicted as surprise. By contrast, fear has a lower recall of 63.60\%, showing that this category remains difficult for the model to identify. Furthermore, on Twitter I and Twitter II, the proposed method shows strong and stable performance for binary emotion classification. On Twitter I, the positive class has a particularly high recall of 97.30\%, while the negative class shows a higher precision of 95.82\%, suggesting that the model effectively detects positive samples while maintaining reliable discrimination for negative samples. On Twitter II, the mean F1-score is 89.03\%, which is slightly lower than that on Twitter I. This performance gap indicates that, although the model generalizes well to binary sentiment classification, it is still affected by differences in data distribution, language usage, and annotation style across datasets. 

In summary, by leveraging both emotional concepts and affective sentences as intermediate semantics, our method helps narrow the "affective gap" and improves IEC performance.

\subsubsection{Comparison with Zero-Shot MLLMs}
\begin{figure}[!t]
  \centering
  \includegraphics[pagebox=mediabox,trim=272.40bp 165.48bp 307.60bp 275.18bp,clip,width=\columnwidth]{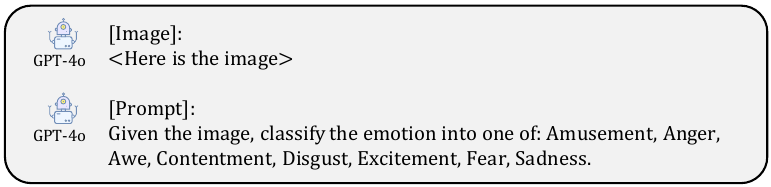} 
  \caption{Zero-shot MLLM prompt template for image emotion classification.}
  \label{fig:zero_shot_mllm_prompt}
\end{figure}

We further compare ACIEC with several zero-shot MLLM baselines, including LLaVA-v1.5 \cite{liu2024llava15}, LLaVA-Next \cite{liu2024llavanext}, Qwen2-VL-7B \cite{wang2024qwen2vl}, and Qwen2.5-VL-7B \cite{qwen2.5vl}, as well as the proprietary model GPT-4o \cite{openai2024gpt4o}. Fig.~\ref{fig:zero_shot_mllm_prompt} presents the prompt template used for these zero-shot baselines. As shown in Table~\ref{tab:VLM}, the evaluated MLLMs obtain lower scores than ACIEC, suggesting that fine-grained emotion discrimination remains challenging for direct zero-shot prediction. ACIEC instead introduces multi-level intermediate 
semantics to help narrow the ``affective gap'' between visual content and emotion labels. However, we emphasize that the reported results are obtained under different learning conditions: ACIEC is trained using task-specific labeled data, whereas the MLLMs are evaluated in a zero-shot setting without adaptation to the target datasets. Therefore, the MLLM results are presented as contextual references for their out-of-the-box performance on fine-grained IEC, rather than as evidence of the intrinsic 
superiority of ACIEC over these models.

In addition, some social media images convey emotion mainly through embedded text, such as overlaid captions, memes, or screenshots. ACIEC explicitly handles such cases with an OCR-based branch, whereas direct zero-shot prompting does not distinguish text-dominant images from text-free ones. This design further improves the robustness of ACIEC in real-world IEC scenarios.

\begin{table}[!t]
\centering
\footnotesize
\setlength{\tabcolsep}{3pt}
\renewcommand{\arraystretch}{1.10}
\renewcommand{\tabularxcolumn}[1]{m{#1}}

\caption{Comparison with zero-shot MLLM baselines on different IEC datasets.}
\label{tab:VLM}
\begin{tabularx}{\linewidth}{@{}>{\raggedright\arraybackslash}X>{\centering\arraybackslash}X>{\centering\arraybackslash}X@{}}
\toprule
\textbf{Model} & \textbf{FI(8)} & \textbf{EmotionROI(6)} \\
\midrule
LLaVA-v1.5 & 53.64 & 49.83 \\
LLaVA-Next & 61.59 & 53.37 \\
Qwen2-VL-7B & 58.16  & 51.68 \\
Qwen2.5-VL-7B & 67.12  & 57.74 \\
GPT-4o & 54.87 & 64.99 \\
\midrule
\textbf{ACIEC} & \textbf{81.98} & \textbf{74.66}  \\
\bottomrule
\end{tabularx}
\end{table}

\subsection{Ablation Studies}
\subsubsection{Component Ablation}
To explore the effectiveness of different components in our method, we compared ACIEC with the following variants. In the first setting, an ANP is used as a single input: "Tem(ANP) = \{[CLS] ANP\}". Similarly, an affective sentence is used as a single input in the second setting: "Tem(DES) = \{[CLS] DES\}". Additionally, we remove the OCR model while keeping both the ANP and the sentence as input. As shown in Table~\ref{Ablation}, omitting either ANP or sentence leads to a clear performance drop, indicating that both emotional concepts and affective sentences contribute to emotion classification and complement each other. Moreover, the affective sentence achieves better performance than the emotional concept on the FI dataset, while the accuracy of the affective sentence is lower than that of the emotional concept on EmotionROI. The reason is that larger datasets cover more diverse image content, for which affective sentences can provide richer and more precise descriptions than the limited set of emotional concepts. In addition, removing the OCR module also reduces performance, which demonstrates that our design effectively filters out noise from images containing embedded text and improves overall accuracy. 

\begin{table*}[!t]
\centering
\small
\setlength{\tabcolsep}{4pt}
\renewcommand{\arraystretch}{1.10}
\renewcommand{\tabularxcolumn}[1]{m{#1}}

\caption{Ablation results of different ACIEC variants on different IEC datasets.}
\label{Ablation}

\begin{tabular*}{\linewidth}{@{\extracolsep{\fill}}lcccccc@{}}
\toprule
\textbf{Method} & \multicolumn{2}{c}{\textbf{FI}} & \multicolumn{2}{c}{\textbf{EmotionROI}} & \textbf{Twitter I} & \textbf{Twitter II} \\
\cmidrule(lr){2-3} \cmidrule(lr){4-5}
\textbf{Number of classes} & \textbf{8} & \textbf{2} & \textbf{6} & \textbf{2} & \textbf{2} & \textbf{2} \\
\midrule
ANP      & 68.48 & 84.17 & 59.25 & 81.32 & 91.07 & 90.83 \\
Sentence & 69.74 & 85.12 & 57.89 & 79.41 & 89.96 & 89.75 \\
w/o OCR      & 78.87 & 88.91 & 69.48 & 85.31 & 91.54 & 89.14 \\
\midrule
Obj           & 63.16 & 72.19 & 50.52 & 72.94 & 45.47 & 44.47 \\
Sce           & 53.48 & 57.52 & 41.47 & 70.66 & 39.78 & 43.06 \\
Obj + Sce     & 69.02 & 82.77 & 54.14 & 75.36 & 54.01 & 59.49 \\
\midrule
Class-level sampling & 78.82 & 93.90 & 72.86 & 89.48 & 91.30 & 89.82 \\
\midrule
\textbf{ACIEC}          & \textbf{81.98} & \textbf{95.68} & \textbf{74.66} & \textbf{92.14} & \textbf{94.54} & \textbf{91.86} \\
\bottomrule
\end{tabular*}
\end{table*}

\subsubsection{Intermediate Semantic Ablation} 
To examine how different intermediate semantics affect the IEC task, we first apply Faster R-CNN \cite{girshick2015fast} to detect objects in the image and retain those with a confidence score of at least 80\%, denoted as "Obj = \{Obj1, Obj2, ...\}". Then, we also extract scene-related words $Sce$ using the method from \cite{zhou2014learning}. A template is constructed as "Tem(Sce, Obj) = \{[CLS] Sce [SEP] Obj\}", and the results are shown in Table~\ref{Ablation}. Additionally, we test inputs using only scene words "Tem(Sce) = \{[CLS] Sce\}" and only object words "Tem(Obj) = \{[CLS] Obj\}". The results show that emotional concepts and affective sentences achieve higher accuracy than object and scene words, demonstrating that they are more strongly related to emotions and thus serve as more suitable intermediate semantics between images and emotions.

\subsubsection{Reasoning Backbone Ablation}

To assess the generalization of the EA-CoT across different reasoning backbones, experiments were conducted not only on the default GPT-4o\cite{openai2024gpt4o} but also on the open-source Qwen2.5-VL-7B\cite{qwen2.5vl} and LLaVA-v1.5\cite{liu2024llava15}. As shown in Table~\ref{backbone}, ACIEC performs consistently across all backbones. Qwen2.5-VL-7B exhibits the most competitive results, performing closely to GPT-4o on most benchmarks. In contrast, LLaVA-v1.5 yields the lowest scores, which can be attributed to its relatively limited capacity for generating affective intermediate semantics. These observations indicate that the performance gains of ACIEC primarily arise from the proposed intermediate semantics rather than from the strong reasoning capability of the LLM. Furthermore, stronger reasoning backbones can improve performance by generating more reliable affective sentences.

\begin{table*}[!t]
\centering
\small
\setlength{\tabcolsep}{4pt}
\renewcommand{\arraystretch}{1.10}
\renewcommand{\tabularxcolumn}[1]{m{#1}}

\caption{Effect of different reasoning backbones on ACIEC.}
\label{backbone}

\begin{tabular*}{\linewidth}{@{\extracolsep{\fill}}lcccccc@{}}
\toprule
\textbf{Method} & \multicolumn{2}{c}{\textbf{FI}} & \multicolumn{2}{c}{\textbf{EmotionROI}} & \textbf{Twitter I} & \textbf{Twitter II} \\
\cmidrule(lr){2-3} \cmidrule(lr){4-5}
\textbf{Number of classes} & \textbf{8} & \textbf{2} & \textbf{6} & \textbf{2} & \textbf{2} & \textbf{2} \\
\midrule
LLaVA-v1.5                    & 79.96 & 94.12 & 71.85 & 90.02 & 92.11 & 90.24 \\
Qwen2.5-VL-7B                   & 80.84 & 94.63 & 72.92 & 90.67 & 92.56 & 90.71 \\
\midrule
\textbf{ACIEC (GPT-4o)}          & \textbf{81.98} & \textbf{95.68} & \textbf{74.66} & \textbf{92.14} & \textbf{94.54} & \textbf{91.86} \\
\bottomrule
\end{tabular*}
\end{table*}

\subsubsection{Loss Function Ablation}
\label{sec:loss-ablation}

\begin{figure*}[!t]
  \centering
  \subfloat[]{%
    \label{confusedmatrix.1}%
    \includegraphics[width=0.46\textwidth]{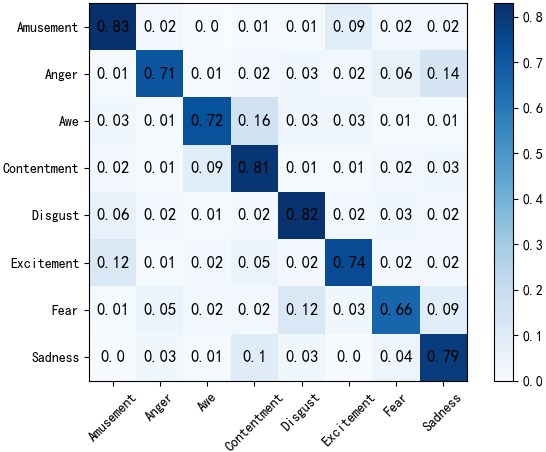}}
  \hfil
  \subfloat[]{%
    \label{confusedmatrix.2}%
    \includegraphics[width=0.46\textwidth]{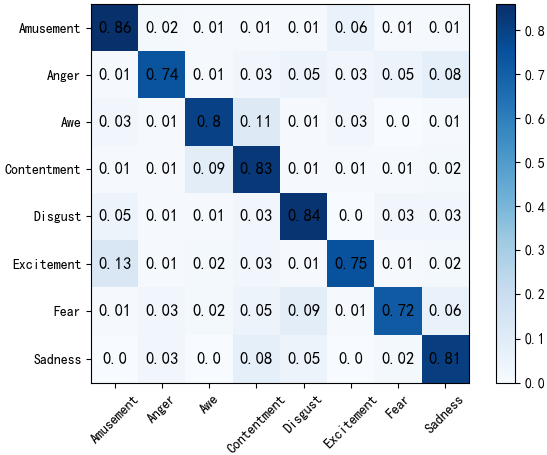}}\\

  \subfloat[]{%
    \label{confusedmatrix.3}%
    \includegraphics[width=0.46\textwidth]{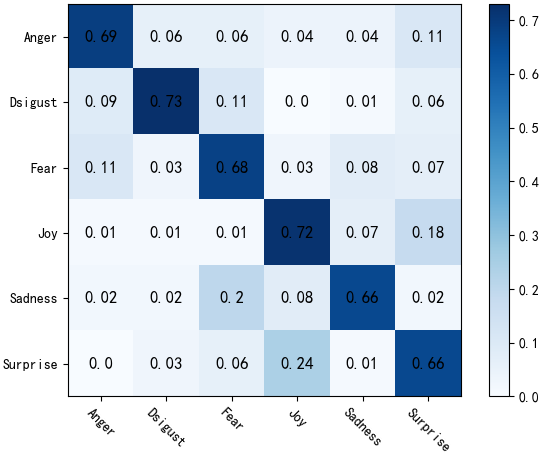}}
  \hfil
  \subfloat[]{%
    \label{confusedmatrix.4}%
    \includegraphics[width=0.46\textwidth]{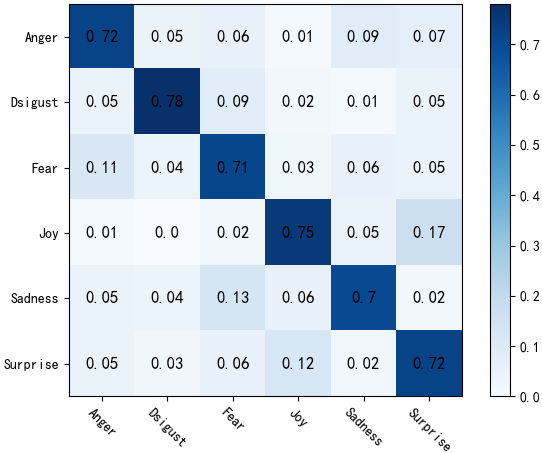}}

  \caption{Confusion matrices for the FI and EmotionROI datasets.}
  \label{confusedmatrix}
\end{figure*}

To validate the effectiveness of the proposed loss, we conduct experiments based on the cross-entropy loss function. The confusion matrices in Fig.~\ref{confusedmatrix} provide an intuitive view of how predictions are distributed across emotion categories. On the FI dataset, the baseline in Fig.~\ref{confusedmatrix.1} exhibits confusion between emotions with the same polarity such as "Amusement" vs. "Excitement", "Anger" vs. "Sadness", and "Awe" vs. "Contentment". With our loss in Fig.~\ref{confusedmatrix.2}, the accuracy of the ``Amusement'' class increases from 83\% to 86\%, and misclassifications between "Amusement" and "Excitement" are reduced by about 3\%, indicating clearer decision boundaries between these highly related emotions. Fig.~\ref{confusedmatrix.3} presents the results of using cross-entropy loss on the EmotionROI dataset. The cross-entropy baseline frequently confuses "anger" with "surprise" and "joy" with "surprise". In contrast, Fig.~\ref{confusedmatrix.4} shows that our loss significantly reduces these errors and improves the per-class accuracy of "disgust", "fear", and "joy". This is because the proposed loss explicitly pulls together multiple ANPs of the same image under the same affective sentence while pushing away ANP--sentence pairs from other categories, thereby enlarging inter-class differences in the embedding space and mitigating confusion among emotions with the same polarity.

Second, we conduct an ablation study on the sampling strategy of our proposed loss function. In the default setting, we take the pair of the top-1 ANP and its corresponding affective sentence of an image as the anchor sample, the pair of the top-2 ANP with the same affective sentence from the same image serves as the positive sample; and pairs of ANPs and affective sentences from the images in different categories are treated as negative samples. This design enables the model to learn consistent and fine-grained emotional representations specific to each image. In contrast, we design a class-level sampling strategy as ablation. In this design, an ANP-sentence pair from a given image still serves as the anchor sample; however, positive samples are now defined as ANP--sentence pairs drawn from a different image that shares the same label, while pairs from images with different labels are treated as negatives. This strategy directly enforces ``pulling together same-class samples and pushing apart different-class samples'' at the label level, without relying on multiple ANPs from the same image. Table~\ref{Ablation} shows the result of the class-level sampling strategy, which indicates that the proposed strategy achieves higher accuracy on fine-grained classes.

Overall, these results demonstrate that our proposed loss effectively alleviates confusion among emotions with the same polarity. Compared with class-level sampling, the proposed semantic similarity-sampled contrastive strategy explicitly enforces semantic consistency among multiple ANPs within the same image while enlarging inter-class margins, thereby making better use of affective intermediate semantics and learning more discriminative emotional representations.

\subsubsection{Analysis of Explicit Emotion Cues}
\label{sec:emotion_cue_analysis}

The EA-CoT-generated affective sentences are intended to provide the descriptions of the affective content of an image. 
However, some affective sentences may directly contain an emotion-label word. Although the EA-CoT prompt provides neither the ground-truth label nor a list of candidate emotion categories, the occurrence of such words could still provide direct lexical cues to the emotion classifier. Consequently, part of the observed performance gain might arise from matching emotion-label words rather than learning richer affective relationships conveyed by the generated descriptions. To examine whether the performance gain arises from richer affective relationships conveyed by the generated descriptions rather than from emotion-label matching, we first measure how often the exact ground-truth class name appears in the affective sentences. We define an explicit label match as a sentence containing the exact word for the image's ground-truth emotion. As reported in Table~\ref{tab:emotion_label_occurrence}, such explicit matches occur in 2,661 of the 20,892 valid affective sentences, corresponding to 12.74\% of the corpus.

\begin{table}[!t]
\centering
\footnotesize
\setlength{\tabcolsep}{3pt}
\renewcommand{\arraystretch}{1.10}
\renewcommand{\tabularxcolumn}[1]{m{#1}}

\caption{Occurrence of exact emotion-label words in affective sentences on FI.}
\label{tab:emotion_label_occurrence}

\begin{tabularx}{\linewidth}{@{}>{\raggedright\arraybackslash}m{0.26\linewidth}>{\centering\hsize=0.8\hsize\linewidth=\hsize\arraybackslash}X>{\centering\hsize=1.2\hsize\linewidth=\hsize\arraybackslash}X@{}}
\toprule
\textbf{Emotion label} & \textbf{Valid sentences} & \textbf{Sentences with emotion label} \\
\midrule
Amusement   & 4,539  & 1,178 (25.95\%) \\
Anger       & 1,143  & 43 (3.76\%)     \\
Awe         & 2,914  & 509 (17.47\%)   \\
Contentment & 5,069  & 368 (7.26\%)    \\
Disgust     & 1,488  & 10 (0.67\%)     \\
Excitement  & 2,490  & 466 (18.71\%)   \\
Fear        & 887    & 53 (5.98\%)     \\
Sadness     & 2,362  & 34 (1.44\%)     \\
\midrule
Total       & 20,892 & 2,661 (12.74\%) \\
\bottomrule
\end{tabularx}
\end{table}

To determine whether ACIEC relies on these direct lexical cues, we conduct an ablation study by removing explicit emotion labels. Specifically, we strip all occurrences of the eight FI class names from every generated affective sentence, regardless of its actual ground-truth category. This filtering is case-insensitive and strictly limited to whole-word matches. By applying this removal lexicon to all samples, we ensure that no ground-truth information is leaked during preprocessing. Finally, we retrain the emotion classifier from scratch using the original data splits, model architecture, and training configurations. As shown in Table~\ref{tab:emotion_label_removal}, completely removing the emotion-label lexicon yields negligible performance drops of only 0.12 and 0.09 percentage points in accuracy (81.98\% to 81.86\%) and Macro-F1 (79.27\% to 79.18\%), respectively. These limited performance changes indicate that ACIEC does not rely substantially on the literal occurrence of the eight FI emotion-category 
names. 

\begin{table}[!t]
\centering
\footnotesize
\setlength{\tabcolsep}{3pt}
\renewcommand{\arraystretch}{1.10}
\renewcommand{\tabularxcolumn}[1]{m{#1}}

\caption{Emotion-label removal ablation on FI.}
\label{tab:emotion_label_removal}

\begin{tabularx}{\linewidth}{@{}>{\raggedright\arraybackslash}m{0.42\linewidth}>{\centering\arraybackslash}X>{\centering\arraybackslash}X@{}}
\toprule
\textbf{Model variant} & \textbf{Accuracy (\%)} & \textbf{Macro-F1 (\%)} \\
\midrule
ACIEC                              & 81.98 & 79.27 \\
ACIEC w/o emotion-label terms      & 81.86 & 79.18 \\
\bottomrule
\end{tabularx}
\end{table}

\subsection{Generalization Evaluation}
\subsubsection{Cross-Dataset Generalization}
\label{sec:cross_dataset}

\begin{table}[!t]
\centering
\footnotesize
\setlength{\tabcolsep}{3pt}
\renewcommand{\arraystretch}{1.10}
\renewcommand{\tabularxcolumn}[1]{m{#1}}

\caption{Cross-dataset generalization results of ACIEC.}
\label{tab:cross_dataset}
\begin{tabularx}{\linewidth}{@{}l>{\centering\arraybackslash}X>{\centering\arraybackslash}X>{\centering\arraybackslash}X>{\centering\arraybackslash}X@{}}
\toprule
\textbf{Test on} & \textbf{Accuracy (\%)} & \textbf{mAP (\%)} & \textbf{m-F1 (\%)} & \textbf{w-F1 (\%)} \\
\midrule
EmotionROI    & 79.84 & 83.96 & 78.98 & 80.39 \\
Twitter I     & 79.38 & 84.06 & 78.64 & 79.63 \\
Twitter II    & 80.47 & 85.28 & 80.39 & 81.25 \\
\bottomrule
\end{tabularx}
\par
\raggedright
\footnotesize \textit{Note:} mAP, m-F1, and w-F1 indicate mean average precision, macro-F1, and weighted-F1, respectively.
\end{table}

To evaluate the robustness of ACIEC to dataset shift, we conduct a cross-dataset generalization evaluation, where FI is used as the source dataset and the remaining datasets are used as target datasets. Following standard protocols in IEC, fine-grained emotion categories are mapped to positive and negative. ACIEC is trained exclusively on FI and directly evaluated on EmotionROI, Twitter I, and Twitter II, without any fine-tuning. As shown in Table~\ref{tab:cross_dataset}, ACIEC achieves consistent performance across all target datasets. The corresponding mAP values remain above 83\%, indicating that the model preserves reliable ranking ability under distribution shift. In addition, the macro-F1 and weighted-F1 scores remain close to the corresponding accuracy values, suggesting that ACIEC retains relatively balanced recognition ability for positive and negative samples. Overall, these results suggest that affective intermediate semantics can help alleviate the "affective gap" and improve cross-dataset generalization under the evaluated setting.

\subsection{Reliability and Efficiency Analysis}
\subsubsection{Confidence Calibration}
\label{sec:calibration}

We further evaluate the confidence calibration of ACIEC using four complementary metrics: Expected Calibration Error (ECE) \cite{guo2017calibration}, Static Calibration Error (SCE) \cite{nixon2019measuring}, Negative Log-Likelihood (NLL), and Brier Score. Specifically, ECE quantifies the misalignment between average confidence and empirical accuracy across 10 equal-width bins, while SCE evaluates class-wise calibration, making it more robust for skewed emotion distributions. NLL measures the probability assigned to the ground-truth label, and the Brier Score computes the mean squared error between the predicted probability distribution and the one-hot label. For all metrics, lower values denote better calibration.

As shown in Table~\ref{tab:calibration}, ACIEC consistently achieves low ECE values across all datasets. Notably, the model yields the best ECE of 0.0258 on Twitter II, alongside a comparable ECE of 0.0265 and the lowest SCE of 0.0213 on the FI dataset, demonstrating highly reliable overall and class-wise calibration. Conversely, EmotionROI exhibits higher ECE, NLL, and Brier Score values compared to FI, suggesting that calibration remains more challenging on smaller, fine-grained affective datasets. On the Twitter datasets, the lower NLL and Brier Scores reflect highly stable probability estimates; however, the relatively higher SCE observed on Twitter I implies that class-wise confidence alignment is still susceptible to data distribution shifts. Overall, these calibration results corroborate our accuracy findings, offering a comprehensive validation of ACIEC under ambiguous and imbalanced emotion distributions.

\begin{table}[!t]
\centering
\footnotesize
\setlength{\tabcolsep}{3pt}
\renewcommand{\arraystretch}{1.10}
\renewcommand{\tabularxcolumn}[1]{m{#1}}

\caption{Confidence calibration results on different IEC datasets.}
\label{tab:calibration}
\begin{tabularx}{\linewidth}{@{}l>{\centering\arraybackslash}X>{\centering\arraybackslash}X>{\centering\arraybackslash}X>{\centering\arraybackslash}X@{}}
\toprule
\textbf{Dataset} & \textbf{ECE} $\downarrow$ & \textbf{SCE} $\downarrow$ & \textbf{NLL} $\downarrow$ & \textbf{Brier Score} $\downarrow$ \\
\midrule
FI            & 0.0265  & 0.0213 & 0.7513 & 0.3566 \\
EmotionROI    & 0.0450  & 0.0381 & 0.7854 & 0.4033 \\
Twitter I     & 0.0366  & 0.0502 & 0.3061 & 0.0911 \\
Twitter II    & 0.0258  & 0.0369 & 0.2673 & 0.0847 \\
\bottomrule
\end{tabularx}
\end{table}

\subsubsection{Statistical Significance Testing}
\label{sec:statistical_significance}

To assess the robustness and statistical reliability of the proposed framework, we conduct five independent runs on the FI dataset under identical experimental settings using different random seeds. The results are shown in Table~\ref{tab:statistical_significance}. StdDev denotes standard deviation. Reported p-values correspond to the paired one-sided t-test and Wilcoxon signed-rank test with the significance level of $\alpha$ = 5\%. It can be observed that our method achieves a mean accuracy of 81.98\% with a standard deviation of 0.14, outperforming the strongest reproducible baseline, which obtains a mean accuracy of 80.63\% with a standard deviation of 0.15.

To further determine whether the observed improvement is statistically significant, we test the null hypothesis $H_{0}$ that ACIEC and the baseline have equal mean performance against the alternative hypothesis $H_{1}$ that ACIEC achieves higher performance. Specifically, we employ both a paired one-sided $t$-test and a Wilcoxon signed-rank test. The paired $t$-test evaluates the significance of the mean performance difference across repeated runs, whereas the Wilcoxon signed-rank test provides a non-parametric validation without assuming normality. The resulting $p$-values are $1.79 \times 10^{-6}$ and $3.13 \times 10^{-2}$, both of which are below the significance threshold of 0.05. Therefore, the null hypothesis is rejected at the 5\% significance level. These results indicate that the performance gain of ACIEC on the FI dataset is statistically significant and is unlikely to be caused by random variation.

\begin{table}[!t]
\centering
\footnotesize
\setlength{\tabcolsep}{3pt}
\renewcommand{\arraystretch}{1.10}
\renewcommand{\tabularxcolumn}[1]{m{#1}}

\caption{Statistical significance testing between ACIEC and the baseline.}
\label{tab:statistical_significance}
\begin{tabularx}{\linewidth}{@{}lcc>{\centering\arraybackslash}X>{\centering\arraybackslash}X@{}}
\toprule
\textbf{Method} & \textbf{Mean (\%)} & \textbf{StdDev} & \textbf{$t$-test $p$} & \textbf{Wilcoxon $p$} \\
\midrule
Baseline & 80.63 & 0.15 & \multirow{2}{*}{$1.79 \times 10^{-6}$} & \multirow{2}{*}{$3.13 \times 10^{-2}$} \\
\textbf{ACIEC} & \textbf{81.98} & 0.14 &  &  \\
\bottomrule
\end{tabularx}
\end{table}

\subsubsection{Computational Cost Analysis}
\label{sec:computational_cost}

\begin{table}[!t]
\centering
\footnotesize
\setlength{\tabcolsep}{3pt}
\renewcommand{\arraystretch}{1.10}
\renewcommand{\tabularxcolumn}[1]{m{#1}}

\caption{Computational cost of the proposed ACIEC framework.}
\label{tab:complexity}
\begin{tabularx}{\linewidth}{@{}l>{\centering\arraybackslash}X>{\centering\arraybackslash}X>{\centering\arraybackslash}X@{}}
\toprule
\textbf{Model} & \textbf{Params (M)} & \textbf{FLOPs (G)} & \textbf{Memory (GB)} \\
\midrule
ACIEC & 85.65 & 21.76 & 11.84 \\
\bottomrule
\end{tabularx}
\end{table}

To further evaluate the efficiency of ACIEC, we analyze its computational cost and resource usage from three aspects: the number of parameters (Params), floating-point operations (FLOPs), and GPU memory footprint (Memory). As reported in Table~\ref{tab:complexity}, ACIEC contains 85.65M parameters, requires 21.76G FLOPs, and consumes 11.84~GB of GPU memory under our implementation configuration. These results indicate that ACIEC is not designed as an extremely lightweight model. Instead, it adopts a moderate-capacity architecture that supports affective semantic modeling while maintaining a practical computational cost for IEC experiments. The primary computational overhead comes from the language-based classification branch, which jointly encodes the detected ANP concepts and the generated affective sentences. Compared with purely visual classifiers, this branch introduces additional parameters and memory consumption. However, it also enables the model to exploit fine-grained affective semantics and to reduce the gap between visual content and emotion labels more effectively. Therefore, the computational cost of ACIEC mainly reflects the representational capacity required for affective semantic reasoning, rather than unnecessary architectural redundancy.

\subsection{Qualitative and Human Evaluation}
\subsubsection{Affective Sentence Quality Analysis}

\begin{table*}[!t]
\centering
\small
\setlength{\tabcolsep}{4pt}
\renewcommand{\arraystretch}{1.10}
\renewcommand{\tabularxcolumn}[1]{m{#1}}

\caption{Qualitative comparison of affective sentences generated by EA-CoT and zero-shot prompting.}
\label{promptcompare}

\begin{tabularx}{\linewidth}{@{}>{\centering\arraybackslash}m{0.12\linewidth}>{\centering\hsize=0.95\hsize\linewidth=\hsize\arraybackslash}X>{\centering\hsize=1.05\hsize\linewidth=\hsize\arraybackslash}X@{}}
\toprule
\textbf{Image} & \textbf{EA-CoT Prompting} & \textbf{Zero-shot Prompting} \\
\midrule
% -------- Row 1 --------
\includegraphics[width=\linewidth,height=2.4cm,keepaspectratio]{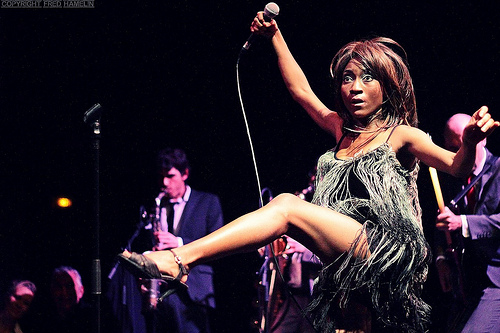}
&
\includegraphics[pagebox=mediabox,trim=539.07bp 261.34bp 195.60bp 155.89bp,clip,width=\linewidth]{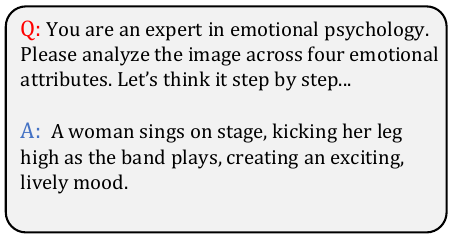}
&
\includegraphics[pagebox=mediabox,trim=288.40bp 144.18bp 435.60bp 274.12bp,clip,width=\linewidth]{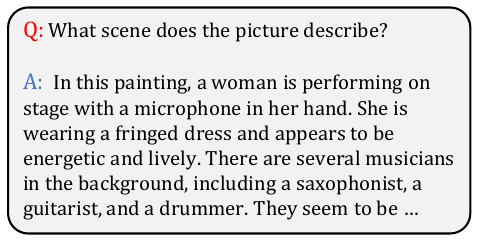}
\\
\midrule
% -------- Row 2 --------
\includegraphics[width=\linewidth,height=2.4cm,keepaspectratio]{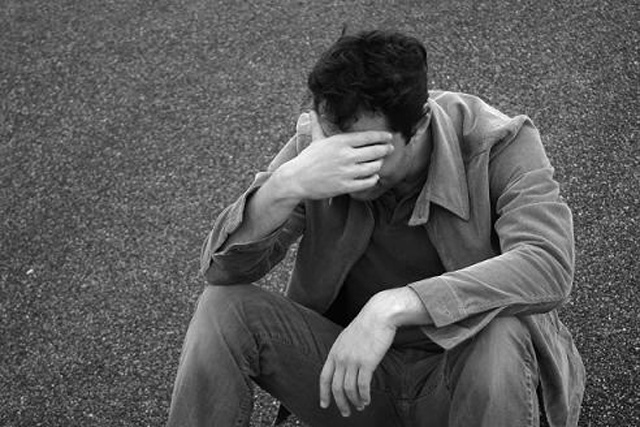}
&
\includegraphics[pagebox=mediabox,trim=538.00bp 144.18bp 195.60bp 274.12bp,clip,width=\linewidth]{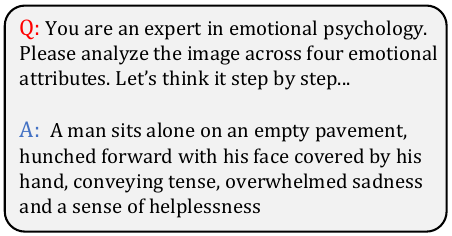}
&
\includegraphics[pagebox=mediabox,trim=288.40bp 275.18bp 435.60bp 143.11bp,clip,width=\linewidth]{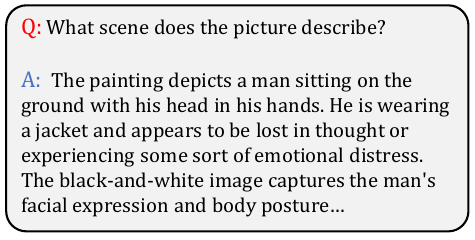}
\\
\bottomrule
\end{tabularx}
\end{table*}

Table~\ref{promptcompare} presents an example comparing the captions for the same image generated by GPT-4o under zero-shot prompting and EA-CoT. With standard zero-shot prompting, the generated caption contains many emotion-irrelevant details, such as listing each artist ("a saxophonist, a guitarist, and a drummer"). These details describe the scene, but they do not highlight the relationship between emotional attributes, which may introduce noise to the affective cues. In contrast, with the guidance of EA-CoT, the generated affective sentence retains the key elements ("sings on stage, kicking her leg high, exciting, lively mood") and focuses on the emotional meaning of the image, enabling better emotion discrimination.

\subsubsection{Human Evaluation of Affective Sentences}
\label{sec:human_evaluation}

\begin{table}[!t]
\centering
\footnotesize
\setlength{\tabcolsep}{3pt}
\renewcommand{\arraystretch}{1.10}
\renewcommand{\tabularxcolumn}[1]{m{#1}}

\caption{Human evaluation results of affective sentences.}
\label{tab:human_eval_mos}
\begin{tabularx}{\linewidth}{@{}l>{\centering\arraybackslash}X>{\centering\arraybackslash}X>{\centering\arraybackslash}X@{}}
\toprule
\textbf{Method} & \textbf{Visual Consistency} & \textbf{Emotional Consistency} & \textbf{Overall MOS} \\
\midrule
Zero-shot      & 3.40$\pm$0.37 & 3.28$\pm$0.35 & 3.34$\pm$0.36 \\
Ours           & 4.10$\pm$0.36 & 4.22$\pm$0.34 & 4.16$\pm$0.35 \\
\bottomrule
\end{tabularx}
\end{table}

To evaluate whether the generated affective sentences align with human perception, we conducted a subjective evaluation on the FI and EmotionROI datasets. We randomly sampled four images per emotion category, yielding 56 images in total (32 from FI, 24 from EmotionROI). For each image, evaluators compared two sentences: one generated via standard zero-shot prompting and the other via our proposed EA-CoT, totaling 112 image-sentence pairs.The study involved 50 evaluators (4 faculty members and 46 students) with sufficient English proficiency and backgrounds in computer vision, cognitive science, or artificial intelligence. After completing written instructions and five practice examples, participants independently rated all 112 pairs. To mitigate source and order biases, generation methods were anonymized, and the two sentences for each image were presented randomly. Ground-truth labels were withheld, ensuring that emotional consistency was judged strictly against the evaluator's own visual perception. Each sentence was rated on a five-point scale along two dimensions: visual consistency (accuracy of salient visual content) and emotional consistency (alignment with the perceived affective impression). The scale ranged from 1 (clear inconsistency) to 5 (high consistency with minor to no limitations). 

The Mean Opinion Score (MOS) for each dimension was averaged across all evaluators, with the overall MOS representing the mean of both dimensions. Inter-rater reliability for the emotional-consistency ratings was assessed using a two-way random-effects, absolute-agreement intraclass correlation coefficient. ICC(2,1) = 0.8534 indicated good agreement among individual evaluators, whereas ICC(2,50) = 0.9966 indicated excellent reliability of the averaged emotional-consistency ratings. As reported in Table~\ref{tab:human_eval_mos}, EA-CoT outperforms zero-shot prompting across all metrics. Visual consistency improves from 3.40 to 4.10, emotional consistency from 3.28 to 4.22, and the overall MOS from 3.34 to 4.16. These results confirm that EA-CoT produces descriptions that are more accurately grounded in visual content and more closely aligned with human affective perception.

% Declare the wide case table early so it stays near its discussion.
\begin{table*}[!t]
\centering
\small
\setlength{\tabcolsep}{4pt}
\renewcommand{\arraystretch}{1.10}
\renewcommand{\tabularxcolumn}[1]{m{#1}}

\caption{Case studies with ANPs and affective sentences.}
\label{casestudy}

\begin{tabularx}{\linewidth}{@{}>{\centering\arraybackslash}m{0.16\linewidth}>{\centering\arraybackslash}m{0.15\linewidth}>{\raggedright\arraybackslash}X>{\centering\arraybackslash}m{0.14\linewidth}@{}}
\toprule
\multicolumn{1}{c}{\textbf{Image}} &
\multicolumn{1}{c}{\textbf{ANP}} &
\multicolumn{1}{c}{\textbf{Affective Sentence}} &
\multicolumn{1}{c}{\textbf{Ground Truth}} \\
\midrule
% Row 1
\includegraphics[width=\linewidth,height=1.5cm,keepaspectratio]{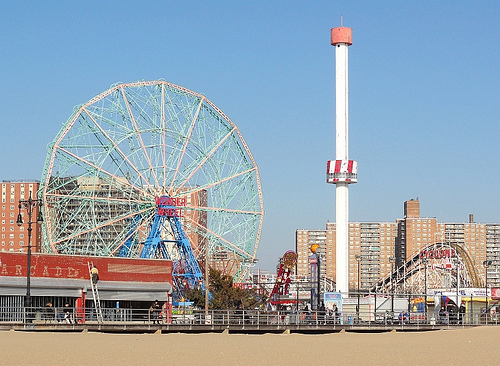}
&
Great adventure
&
A distant view of a beachfront amusement area with a large Ferris wheel and a tall tower ride, set against city buildings under a clear sky.
&
Amusement
\\
\midrule
% Row 2
\includegraphics[width=\linewidth,height=1.5cm,keepaspectratio]{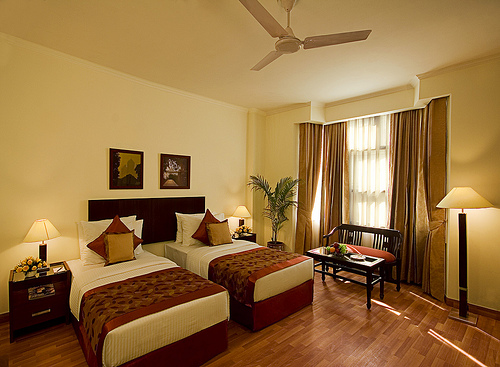}
&
Comfortable room
&
A warm-lit bedroom with two single beds, matching bedside lamps, and a curtained window, with a small table and chair set near the light.
&
Contentment
\\
\midrule
% Row 3
\includegraphics[width=\linewidth,height=1.5cm,keepaspectratio]{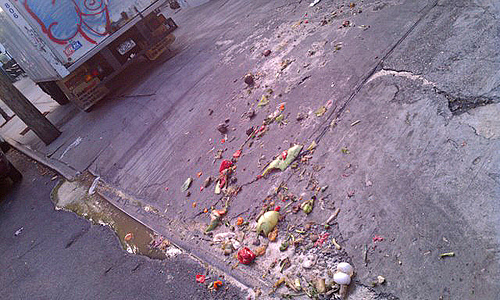}
&
Dirty streets
&
Food scraps and litter are scattered across cracked pavement near a curb and a parked truck, creating a messy street-corner scene.
&
Disgust
\\
\midrule
% Row 4
\includegraphics[width=\linewidth,height=1.5cm,keepaspectratio]{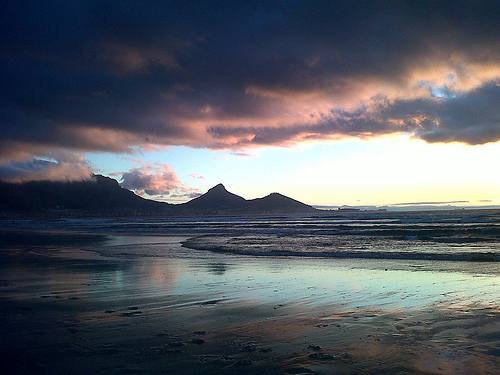}
&
Heavy clouds
&
A quiet shoreline reflecting the sunset, with dark clouds and distant mountains, calling forth serene reflection.
&
Awe
\\
\midrule
% Row 5
\includegraphics[width=\linewidth,height=1.5cm,keepaspectratio]{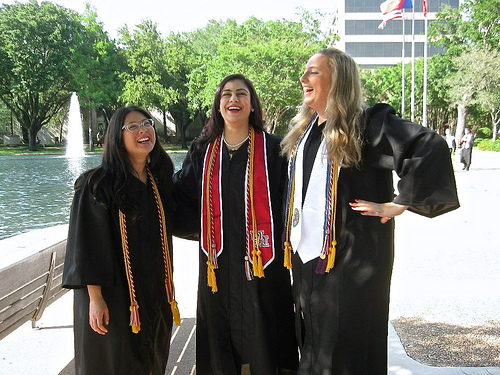}
&
Talented student
&
Three graduates laugh together outdoors in their gowns and stoles, conveying joyful celebration in a milestone moment.
&
Excitement
\\
\bottomrule
\end{tabularx}
\end{table*}

\subsubsection{Feature-Space Visualization}
\label{ssec:feature_space_visualization}

\begin{figure*}[!t]
  \centering 
  \subfloat[]{  
    \includegraphics[width=0.48\textwidth]{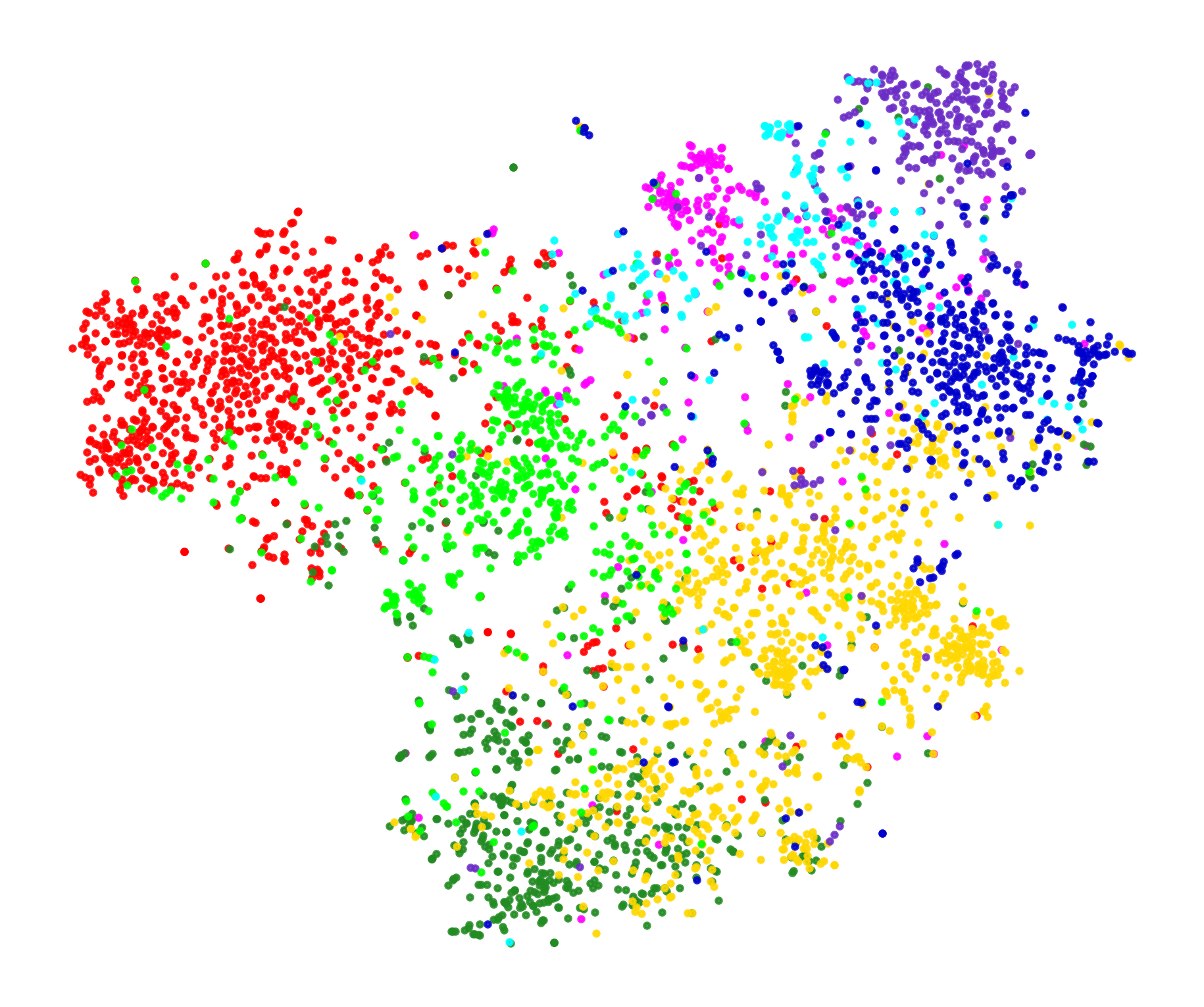}}\hfil 
  \subfloat[]{ 
    \includegraphics[width=0.48\textwidth]{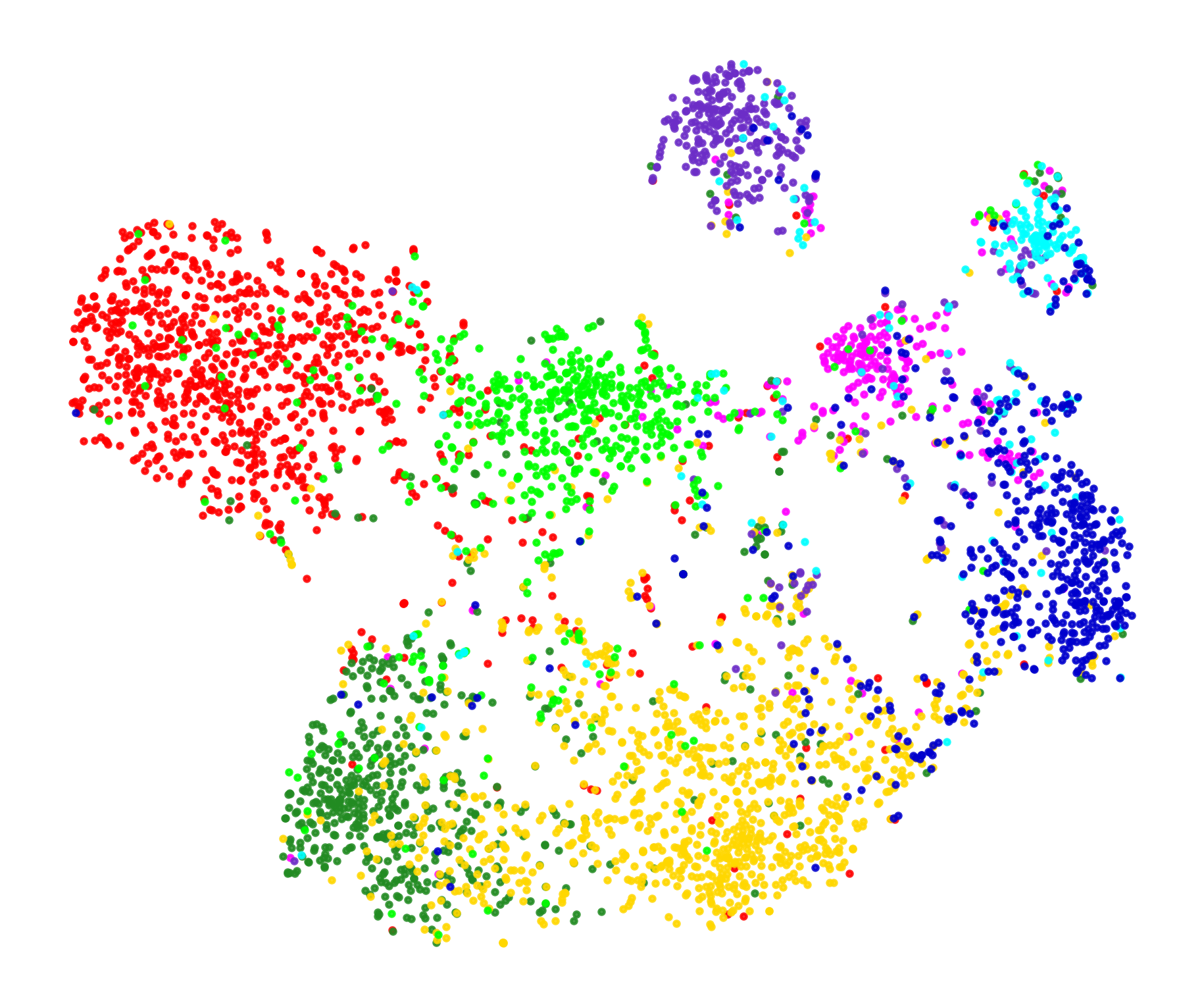}} 
  \caption{t-SNE visualization of intermediate semantic features.} 
  \label{tsne}
\end{figure*}

To provide an intuitive illustration of the effectiveness of our method, we visualize the feature vectors learned from our model with t-distributed Stochastic Neighbor Embedding (t-SNE) in Fig.~\ref{tsne}. Each point represents an image feature vector in the test set (3316 images for FI) and different colors denote different emotion categories. It can be seen that in the first plot, the points are distributed in several clusters. Inter-class clusters are clearly separated, and intra-class samples are relatively compact. These results demonstrate that a highly discriminative embedding space is produced with the guidance of both emotional concepts and affective sentences. In contrast, the second plot shows the results using only affective sentences as the final semantic representation. Here, more overlap between classes is observed, indicating that affective sentences lack the abstractness offered by emotional concepts, making it more difficult for the model to distinguish similar emotions. The difference between the two plots highlights the complementary nature of emotional concepts and affective sentences. Emotional concepts help the model understand the broader emotional context, while affective sentences provide specific details.

% Queue the final tables before their discussion, preserving table order.
\begin{table}[!t]
\centering
\footnotesize
\setlength{\tabcolsep}{3pt}
\renewcommand{\arraystretch}{1.10}
\renewcommand{\tabularxcolumn}[1]{m{#1}}

\caption{Performance on abstract and artistic emotion datasets.}
\label{tab:results_abstract_artphoto}
\begin{tabularx}{\linewidth}{@{}l>{\centering\arraybackslash}X>{\centering\arraybackslash}X>{\centering\arraybackslash}X>{\centering\arraybackslash}X@{}}
\toprule
\textbf{Dataset} & \textbf{Accuracy (\%)} & \textbf{mAP (\%)} & \textbf{m-F1 (\%)} & \textbf{W-F1 (\%)} \\
\midrule
Abstract & 48.13 & 49.99 & 44.43 & 45.91 \\
ArtPhoto & 31.57 & 28.41 & 26.23 & 27.60 \\
\bottomrule
\end{tabularx}
\par
\raggedright
\footnotesize \textit{Note:} mAP, m-F1, and w-F1 indicate mean average precision, macro-F1, and weighted-F1, respectively.
\end{table}

\begin{table*}[!t]
\centering
\small
\setlength{\tabcolsep}{4pt}
\renewcommand{\arraystretch}{1.10}
\renewcommand{\tabularxcolumn}[1]{m{#1}}

\caption{Failure cases on abstract and artistic emotion datasets.}
\label{tab:anp_examples}

\begin{tabularx}{\linewidth}{@{}>{\centering\arraybackslash}m{0.12\linewidth}>{\centering\arraybackslash}m{0.16\linewidth}>{\centering\arraybackslash}m{0.17\linewidth}>{\raggedright\arraybackslash}X@{}}
\toprule
\multicolumn{1}{c}{\textbf{Dataset}} &
\multicolumn{1}{c}{\textbf{Image}} &
\multicolumn{1}{c}{\textbf{ANP}} &
\multicolumn{1}{c}{\textbf{Affective Sentence}} \\
\midrule

% Row 1
\multirow{2}{*}[-20pt]{ArtPhoto} & 
\includegraphics[width=\linewidth,height=1.5cm,keepaspectratio]{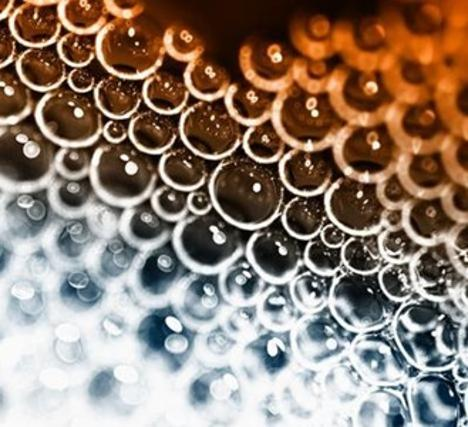} & 
bumpy water & 
The bubbles catch the light in an interesting way, making the scene feel curious and a little magical. \\
\cmidrule{2-4}
% Row 2
& 
\includegraphics[width=\linewidth,height=1.5cm,keepaspectratio]{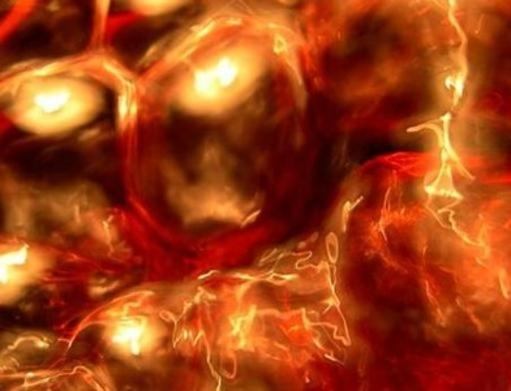} & 
bloody hands & 
The fiery colors and rough texture make the image feel full of energy and strong emotion. \\
\midrule

% Row 3
\multirow{2}{*}[-20pt]{Abstract} & 
\includegraphics[width=\linewidth,height=1.5cm,keepaspectratio]{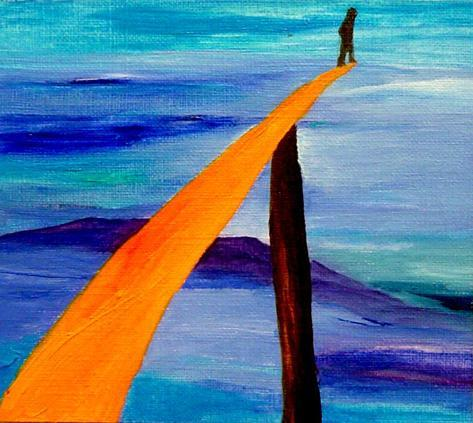} & 
lonely road & 
Solitary journey on a vibrant path, bridging uncertainty with hope amidst an expansive, introspective horizon. \\
\cmidrule{2-4}
% Row 4
& 
\includegraphics[width=\linewidth,height=1.5cm,keepaspectratio]{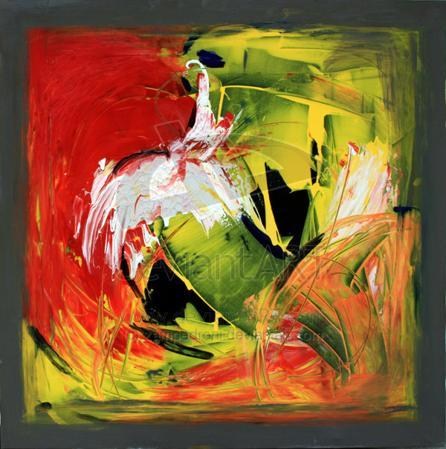} & 
amazing paintings & 
An abstract dance of fiery reds and vibrant greens, where bold textures and dynamic strokes evoke passion, vitality, and the chaos of raw emotion. \\
\bottomrule
\end{tabularx}
\end{table*}

\subsection{Case study}
\label{ssec:case study}

To provide a clearer understanding of the proposed method, we visualize the examples of images with both emotional concepts and affective sentences. As shown in Table~\ref{casestudy}, in first three images, ANPs such as ``great adventure,'' ``comfortable room,'' and ``dirty streets'' offer concise, highly abstract summaries that closely match the visual content and directly express the emotion. The corresponding sentences describe objects, scenes, and actions in greater detail, but their connection to affect is more implicit and thus less aligned with the target emotion. In the last two images, the ANPs partially deviate from the actual image semantics, indicating the limited size of the ANP vocabulary and the coverage constraints of the VSO dataset, which restrict descriptive accuracy. In these cases, the affective sentences provide a more reliable inference of emotion. Table~\ref{casestudy} highlights the complementarity between emotional concepts and affective sentences.

\subsection{Limitations}
\label{sec:limitations}

Although ACIEC provides a new way to bridge visual content and emotional semantics and achieves favorable results on several IEC benchmarks, it still has several limitations. First, the ANP detector is built on the VSO dataset, and therefore its vocabulary and domain coverage are inherently constrained by the scale and diversity of VSO. When the affective content of an image cannot be well matched to existing ANP categories, the detected ANPs may become incomplete or inaccurate. ACIEC alleviates this limitation by combining emotional concepts with EA-CoT-generated affective sentences. However, this design does not fully eliminate the vocabulary bottleneck of ANPs. Future work will explore open-vocabulary emotional concept discovery and domain-adaptive ANP expansion.

Second, the proposed framework is primarily designed for images that contain clear semantic objects and recognizable visual cues. Certain image types, such as abstract paintings and highly artistic images, are difficult to represent with ANPs because they lack well-defined objects or object-centric structures. To examine this limitation, we further evaluate ACIEC on two artistic datasets. Table~\ref{tab:results_abstract_artphoto} shows the performance results of our method on ArtPhoto and Abstract Paintings \cite{machajdik2010affective}. It can be observed that the performance of the proposed method decreases substantially on both datasets. This observation is consistent with the design of ACIEC, which relies on meaningful intermediate textual representations generated from recognizable visual cues. When such cues are partially available, as in many ArtPhoto images, the model can still extract useful semantic information to some extent, although the resulting performance is clearly lower than that on conventional social-media affective datasets. In contrast, for highly abstract or non-object-centric images, explicit semantic anchors are often absent, making it difficult for the ANP detector and the caption generator to produce semantically aligned textual representations.

To make this limitation more transparent, we further include case studies from both the ArtPhoto and Abstract Paintings datasets. As shown in Table~\ref{tab:anp_examples}, the generated ANPs and affective sentences for these datasets are only weakly related to the corresponding images. For example, outputs such as ``bloody hands'' and ``lonely road'' capture only surface-level visual impressions and fail to establish a stable correspondence with the underlying emotion categories. These examples further confirm that the proposed framework is designed primarily for images with recognizable semantic content, whereas its effectiveness is limited for highly artistic, abstract, or non-object-centric imagery.

Furthermore, biases in datasets and annotations represent another important limitation. The experimental datasets are primarily collected from social media platforms and annotated through crowd sourcing protocols. Consequently, the emotion labels may reflect cultural, demographic, temporal, and annotator-specific biases. Because these benchmarks do not provide reliable demographic, cultural, geographic, or temporal metadata, rigorous subgroup-level fairness analysis cannot be performed without introducing additional assumptions. Future work should construct or adopt IEC datasets with richer metadata and fairness-oriented annotation protocols to enable systematic analysis of cultural, demographic, and temporal biases.

% Keep all experimental figures and tables before the conclusion.
\FloatBarrier
\section{Conclusion}
\label{sec:conclusion}
In this paper, we leverage multi-level image captions as intermediate semantics to help mitigate the "affective gap" in image emotion classification. We introduce a hierarchical multi-level contrastive loss and an emotional-attribute chain-of-thought prompt to generate emotional concepts and affective sentences, respectively. To address the challenge of large intra-class variation and small inter-class margins commonly observed in affective datasets, we further propose a semantic-similarity sampling--based contrastive learning loss. Moreover, considering that the presence of text in social media images may introduce noise, we incorporate an OCR module to detect and handle such cases, thereby improving the robustness of the model. Comprehensive experiments on multiple affective benchmarks demonstrate the effectiveness of the proposed approach.

\section*{Declarations}
\subsection*{Ethics approval and consent to participate}
Not applicable.
\subsection*{Availability of data and material}
The datasets used and/or analyzed during the current study available from the corresponding author on reasonable request.
\subsection*{Competing interests}
The authors have no competing interests to declare that are relevant to the content of this article.
\subsection*{Author contribution}
Conceptualization: ZZ (Zhengjun Zhai) and ZZ (Zibo Zhou); methodology: ZZ (Zibo Zhou); Formal analysis and investigation: ZZ (Zibo Zhou); validation: ZZ (Zibo Zhou) and HC; writing-original draft preparation: HC; writing-review and editing: ZZ (Zibo Zhou) and HC. Supervision: ZZ (Zhengjun Zhai) All authors have read and agreed to the published version of the manuscript. 
\subsection*{Funding}
Not applicable.

\bibliographystyle{IEEEtran}
\bibliography{references}

@inproceedings{siersdorfer2010analyzing,
  author    = {Siersdorfer, Stefan and Minack, Enrico and Deng, Fan and Hare, Jon},
  title     = {Analyzing and Predicting Sentiment of Images on the Social Web},
  booktitle = {Proceedings of the 18th ACM International Conference on Multimedia (ACM MM)},
  year      = {2010},
  pages     = {715--718}
}

@inproceedings{zhao2014exploring,
  author    = {Zhao, Sicheng and Gao, Y. and Jiang, X. and Yao, Hongxun and Chua, Tat-Seng and Sun, X.},
  title     = {Exploring Principles-of-Art Features for Image Emotion Recognition},
  booktitle = {Proceedings of the 22nd ACM International Conference on Multimedia (ACM MM)},
  year      = {2014},
  pages     = {47--56}
}

@book{mehrabian1974approach,
  title={An approach to environmental psychology.},
  author={Mehrabian, Albert and Russell, James A},
  year={1974},
  publisher={the MIT Press}
}

@article{yang2021solver,
  author  = {Yang, J. and Gao, X. and Li, L. and Wang, X. and Ding, J.},
  title   = {Solver: Scene-object Interrelated Visual Emotion Reasoning Network},
  journal = {IEEE Transactions on Image Processing},
  volume  = {30},
  pages   = {8686--8701},
  year    = {2021}
}

@article{cen2024masanet,
  author  = {Cen, J. and Qing, C. and Ou, H. and Xu, X. and Tan, J.},
  title   = {Masanet: Multi-aspect Semantic Auxiliary Network for Visual Sentiment Analysis},
  journal = {IEEE Transactions on Affective Computing},
  volume  = {15},
  number  = {3},
  pages   = {1439--1450},
  year    = {2024}
}

@inproceedings{deng2022simemotion,
  author    = {Deng, S. and Shi, G. and Wu, L. and Xing, L. and Hu, W. and Zhang, H. and Xiang, Y.},
  title     = {Simemotion: A Simple Knowledgeable Prompt Tuning Method for Image Emotion Classification},
  booktitle = {Proceedings of the International Conference on Database Systems for Advanced Applications (DASFAA)},
  year      = {2022},
  pages     = {222--229}
}

@inproceedings{borth2013large,
  author    = {Borth, Damian and Ji, Rongrong and Chen, Tao and Breuel, Thomas and Chang, Shih-Fu},
  title     = {Large-scale Visual Sentiment Ontology and Detectors Using Adjective Noun Pairs},
  booktitle = {Proceedings of the 21st ACM International Conference on Multimedia (ACM MM)},
  year      = {2013},
  pages     = {223--232}
}

@inproceedings{lu2012shape,
  author    = {Lu, Xin and Suryanarayan, P. and Adams Jr., Reginald B. and Li, J. and Newman, Michael G. and Wang, James Z.},
  title     = {On Shape and the Computability of Emotions},
  booktitle = {Proceedings of the 20th ACM International Conference on Multimedia (ACM MM)},
  year      = {2012},
  pages     = {229--238}
}

@inproceedings{zhan2019zero,
  author    = {Zhan, Chunyan and She, Dongyu and Zhao, Sicheng and Cheng, Ming-Ming and Yang, Jian},
  title     = {Zero-shot Emotion Recognition via Affective Structural Embedding},
  booktitle = {Proceedings of the IEEE/CVF International Conference on Computer Vision (ICCV)},
  year      = {2019},
  pages     = {1151--1160}
}

@article{zhang2020object,
  author  = {Zhang, J. and Chen, M. and Sun, H. and Li, D. and Wang, Z.},
  title   = {Object Semantics Sentiment Correlation Analysis Enhanced Image Sentiment Classification},
  journal = {Knowledge-Based Systems},
  volume  = {191},
  pages   = {105245},
  year    = {2020}
}

@inproceedings{lin2020multi,
  author    = {Lin, C. and Zhao, Sicheng and Meng, L. and Chua, Tat-Seng},
  title     = {Multi-source Domain Adaptation for Visual Sentiment Classification},
  booktitle = {Proceedings of the AAAI Conference on Artificial Intelligence},
  volume    = {34},
  number    = {3},
  pages     = {2661--2668},
  year      = {2020}
}

@inproceedings{zhu2025learning,
  author    = {Zhu, J. and Zhao, Sicheng and Jiang, J. and Xu, Z. and Tang, W. and Yao, H.},
  title     = {Learning Class Prototypes for Visual Emotion Recognition},
  booktitle = {Proceedings of the IEEE International Conference on Acoustics, Speech and Signal Processing (ICASSP)},
  year      = {2025},
  pages     = {1--5}
}

@article{zhou2025object,
  title={Object-scene semantics correlation analysis for image emotion classification},
  author={Zhou, Zibo and Zhai, Zhengjun and Chen, Huimin and Lu, Sheng},
  journal={Frontiers in Neuroscience},
  volume={19},
  pages={1657562},
  year={2025},
  publisher={Frontiers Media SA}
}

@article{wei2022chain,
  author  = {Wei, Jason and Wang, X. and Schuurmans, Dale and Bosma, Maarten and Xia, Fei and Chi, Ed and Le, Quoc V. and Zhou, Denny and others},
  title   = {Chain-of-thought Prompting Elicits Reasoning in Large Language Models},
  journal = {Advances in Neural Information Processing Systems},
  volume  = {35},
  pages   = {24824--24837},
  year    = {2022}
}

@article{wang2025multimodal,
  author  = {Wang, Y. and Wu, S. and Zhang, Y. and Yan, S. and Liu, Z. and Luo, J. and Fei, H.},
  title   = {Multimodal Chain-of-thought Reasoning: A Comprehensive Survey},
  journal = {arXiv preprint},
  volume  = {arXiv:2503.12605},
  year    = {2025}
}

@article{kojima2022large,
  author  = {Kojima, Takeshi and Gu, Shixiang Shane and Reid, M. and Matsuo, Yutaka and Iwasawa, Yusuke},
  title   = {Large Language Models are Zero-shot Reasoners},
  journal = {Advances in Neural Information Processing Systems},
  volume  = {35},
  pages   = {22199--22213},
  year    = {2022}
}

@article{wu2024enhancing,
  author  = {Wu, J. and Shen, Y. and Zhang, Z. and Cai, L.},
  title   = {Enhancing Large Language Model with Decomposed Reasoning for Emotion Cause Pair Extraction},
  journal = {arXiv preprint},
  volume  = {arXiv:2401.17716},
  year    = {2024}
}

@article{li2024enhancing,
  author  = {Li, Z. and Chen, G. and Shao, R. and Xie, Y. and Jiang, D. and Nie, L.},
  title   = {Enhancing Emotional Generation Capability of Large Language Models via Emotional Chain-of-thought},
  journal = {arXiv preprint},
  volume  = {arXiv:2401.06836},
  year    = {2024}
}

@inproceedings{he2020momentum,
  author    = {He, Kaiming and Fan, Haoqi and Wu, Yuxin and Xie, Saining and Girshick, Ross},
  title     = {Momentum Contrast for Unsupervised Visual Representation Learning},
  booktitle = {Proceedings of the IEEE/CVF Conference on Computer Vision and Pattern Recognition (CVPR)},
  year      = {2020},
  pages     = {9729--9738}
}

@inproceedings{chen2020simple,
  author    = {Chen, Ting and Kornblith, Simon and Norouzi, Mohammad and Hinton, Geoffrey},
  title     = {A Simple Framework for Contrastive Learning of Visual Representations},
  booktitle = {Proceedings of the International Conference on Machine Learning (ICML)},
  year      = {2020},
  pages     = {1597--1607}
}

@article{fan2021unsupervised,
  author  = {Fan, Haoqi and Liu, Ping and Xu, Ming and Yang, Yi},
  title   = {Unsupervised Visual Representation Learning via Dual-level Progressive Similar Instance Selection},
  journal = {IEEE Transactions on Cybernetics},
  volume  = {52},
  number  = {9},
  pages   = {8851--8861},
  year    = {2021}
}

@inproceedings{zhong2021neighborhood,
  author    = {Zhong, Zhun and Fini, Enrico and Roy, Subhankar and Luo, Z. and Ricci, Elisa and Sebe, Nicu},
  title     = {Neighborhood Contrastive Learning for Novel Class Discovery},
  booktitle = {Proceedings of the IEEE/CVF Conference on Computer Vision and Pattern Recognition (CVPR)},
  year      = {2021},
  pages     = {10867--10875}
}

@article{chen2014deepsentibank,
  author  = {Chen, Tao and Borth, Damian and Darrell, Trevor and Chang, Shih-Fu},
  title   = {{DeepSentibank}: Visual Sentiment Concept Classification with Deep Convolutional Neural Networks},
  journal = {arXiv preprint},
  volume  = {arXiv:1410.8586},
  year    = {2014}
}

@inproceedings{jia2014caffe,
  author    = {Jia, Yangqing and Shelhamer, Evan and Donahue, Jeff and Karayev, Sergey and Long, Jonathan and Girshick, Ross and Guadarrama, Sergio and Darrell, Trevor},
  title     = {Caffe: Convolutional Architecture for Fast Feature Embedding},
  booktitle = {Proceedings of the 22nd ACM International Conference on Multimedia (ACM MM)},
  year      = {2014},
  pages     = {675--678}
}

@inproceedings{radford2021learning,
  author    = {Radford, Alec and Kim, Jong Wook and Hallacy, Chris and Ramesh, Aditya and Goh, Gabriel and Agarwal, Sandhini and Sastry, Girish and Askell, Amanda and Mishkin, Pamela and Clark, Jack and others},
  title     = {Learning Transferable Visual Models from Natural Language Supervision},
  booktitle = {Proceedings of the International Conference on Machine Learning (ICML)},
  year      = {2021},
  pages     = {8748--8763}
}

@article{brosch2010perception,
  author  = {Brosch, Tobias and Pourtois, Gilles and Sander, David},
  title   = {The Perception and Categorisation of Emotional Stimuli: A Review},
  journal = {Cognition and Emotion},
  volume  = {24},
  number  = {3},
  pages   = {377--400},
  year    = {2010}
}

@article{steward2025interactions,
  author  = {Steward, B. A. and Mewton, P. and Palermo, R. and Dawel, A.},
  title   = {Interactions between Faces and Visual Context in Emotion Perception: A Meta-analysis},
  journal = {Psychonomic Bulletin \& Review},
  year    = {2025},
  pages   = {1--17},
  note    = {pages/volume may vary by final publication}
}

@article{ekman1993facial,
  author  = {Ekman, Paul},
  title   = {Facial Expression and Emotion},
  journal = {American Psychologist},
  volume  = {48},
  number  = {4},
  pages   = {384},
  year    = {1993}
}

@article{de2015perception,
  author  = {de Gelder, Beatrice and de Borst, Anouk W. and Watson, Robert},
  title   = {The perception of emotion in body expressions},
  journal = {Wiley Interdisciplinary Reviews: Cognitive Science},
  volume  = {6},
  number  = {2},
  pages   = {149--158},
  year    = {2015},
  doi     = {10.1002/wcs.1335}
}

@article{frijda2009emotion,
  author  = {Frijda, Nico H.},
  title   = {Emotion Experience and Its Varieties},
  journal = {Emotion Review},
  volume  = {1},
  number  = {3},
  pages   = {264--271},
  year    = {2009}
}

@article{bar2004visual,
  author  = {Bar, Moshe},
  title   = {Visual Objects in Context},
  journal = {Nature Reviews Neuroscience},
  volume  = {5},
  number  = {8},
  pages   = {617--629},
  year    = {2004}
}

@article{wang2022self,
  author  = {Wang, X. and Wei, J. and Schuurmans, D. and Le, Q. and Chi, E. and Narang, S. and Chowdhery, A. and Zhou, D.},
  title   = {Self-consistency Improves Chain of Thought Reasoning in Language Models},
  journal = {arXiv preprint},
  volume  = {arXiv:2203.11171},
  year    = {2022}
}

@article{cui2025paddleocr,
  author  = {Cui, C. and Sun, T. and Lin, M. and Gao, T. and Zhang, Y. and Liu, J. and Wang, X. and Zhang, Z. and Zhou, C. and Liu, H. and others},
  title   = {{PaddleOCR} 3.0 Technical Report},
  journal = {arXiv preprint},
  volume  = {arXiv:2507.05595},
  year    = {2025}
}

@inproceedings{devlin2019bert,
  title={{BERT}: Pre-training of Deep Bidirectional Transformers for Language Understanding},
  author={Devlin, Jacob and Chang, Ming-Wei and Lee, Kenton and Toutanova, Kristina},
  booktitle={Proceedings of the 2019 Conference of the North American Chapter of the Association for Computational Linguistics: Human Language Technologies, Volume 1 (Long and Short Papers)},
  pages={4171--4186},
  year={2019}
}

@inproceedings{you2015robust,
  author    = {You, Quanzeng and Luo, Jiebo and Jin, Hailin and Yang, Jianchao},
  title     = {Robust Image Sentiment Analysis Using Progressively Trained and Domain Transferred Deep Networks},
  booktitle = {Proceedings of the AAAI Conference on Artificial Intelligence},
  volume    = {29},
  number    = {1},
  year      = {2015}
}

@inproceedings{peng2015mixed,
  title={A mixed bag of emotions: Model, predict, and transfer emotion distributions},
  author={Peng, Kuan-Chuan and Chen, Tsuhan and Sadovnik, Amir and Gallagher, Andrew C},
  booktitle={Proceedings of the IEEE conference on computer vision and pattern recognition},
  pages={860--868},
  year={2015}
}

@inproceedings{you2016building,
  author    = {You, Quanzeng and Luo, Jiebo and Jin, Hailin and Yang, Jianchao},
  title     = {Building a Large Scale Dataset for Image Emotion Recognition: The Fine Print and the Benchmark},
  booktitle = {Proceedings of the AAAI Conference on Artificial Intelligence},
  volume    = {30},
  number    = {1},
  year      = {2016}
}

@article{mikels2005emotional,
  author    = {Mikels, Joseph A. and Fredrickson, Barbara L. and Larkin, Gregory R. and Lindberg, Casey M. and Maglio, Sam J. and Reuter-Lorenz, Patricia A.},
  title     = {Emotional Category Data on Images from the International Affective Picture System},
  journal   = {Behavior Research Methods},
  volume    = {37},
  number    = {4},
  pages     = {626--630},
  year      = {2005},
  publisher = {Springer},
  doi       = {10.3758/BF03192732}
}

@article{liu2019roberta,
  author  = {Liu, Yinhan and Ott, Myle and Goyal, Naman and Du, Jingfei and Joshi, Mandar and Chen, Danqi and Levy, Omer and Lewis, Mike and Zettlemoyer, Luke and Stoyanov, Veselin},
  title   = {{RoBERTa}: A Robustly Optimized {BERT} Pretraining Approach},
  journal = {arXiv preprint},
  volume  = {arXiv:1907.11692},
  year    = {2019}
}

@article{she2019wscnet,
  author  = {She, Dongyu and Yang, Jufeng and Cheng, Ming-Ming and Lai, Yu-Kun and Rosin, Paul L. and Wang, Liang},
  title   = {{WSCNet}: Weakly Supervised Coupled Networks for Visual Sentiment Classification and Detection},
  journal = {IEEE Transactions on Multimedia},
  volume  = {22},
  number  = {5},
  pages   = {1358--1371},
  year    = {2020},
  doi     = {10.1109/TMM.2019.2939744}
}

@article{zhang2022image,
  author  = {Zhang, J. and Liu, X. and Chen, M. and Ye, Q. and Wang, Z.},
  title   = {Image Sentiment Classification via Multi-level Sentiment Region Correlation Analysis},
  journal = {Neurocomputing},
  volume  = {469},
  pages   = {221--233},
  year    = {2022}
}

@article{zhang2024object,
  author  = {Zhang, J. and Liu, J. and Ding, W. and Wang, Z.},
  title   = {Object Aroused Emotion Analysis Network for Image Sentiment Analysis},
  journal = {Knowledge-Based Systems},
  volume  = {286},
  pages   = {111429},
  year    = {2024}
}

@article{shi2023one,
  author  = {Shi, G. and Deng, S. and Wang, B. and Feng, C. and Zhuang, Y. and Wang, X.},
  title   = {One for All: A Unified Generative Framework for Image Emotion Classification},
  journal = {IEEE Transactions on Circuits and Systems for Video Technology},
  volume  = {34},
  number  = {8},
  pages   = {7057--7068},
  year    = {2023}
}

@article{yang2024concept,
  author  = {Yang, H. and Fan, Y. and Lv, G. and Liu, S. and Guo, Z.},
  title   = {Concept-guided Multi-level Attention Network for Image Emotion Recognition},
  journal = {Signal, Image and Video Processing},
  volume  = {18},
  number  = {5},
  pages   = {4313--4326},
  year    = {2024}
}

@article{chen2025dynamic,
  title={Dynamic multi-scenario prompt learning with knowledge augmentation for image emotion analysis},
  author={Chen, Ting and Zhang, Guan and Wei, Yujie and Chen, Jun and Feng, Chen},
  journal={Image and Vision Computing},
  pages={105884},
  year={2025}
}

@article{zhou2022learning,
  author  = {Zhou, Kaiyang and Yang, Jingkang and Loy, Chen Change and Liu, Ziwei},
  title   = {Learning to Prompt for Vision-Language Models},
  journal = {International Journal of Computer Vision},
  volume  = {130},
  number  = {9},
  pages   = {2337--2348},
  year    = {2022},
  doi     = {10.1007/s11263-022-01653-1}
}

@article{deng2024learning,
  author  = {Deng, S. and Wu, L. and Shi, G. and Xing, L. and Jian, M. and Xiang, Y. and Dong, R.},
  title   = {Learning to Compose Diversified Prompts for Image Emotion Classification},
  journal = {Computational Visual Media},
  volume  = {10},
  number  = {6},
  pages   = {1169--1183},
  year    = {2024}
}

@inproceedings{liu2024llava15,
  author    = {Liu, Haotian and Li, Chunyuan and Li, Yuheng and Lee, Yong Jae},
  title     = {Improved Baselines with Visual Instruction Tuning},
  booktitle = {Proceedings of the IEEE/CVF Conference on Computer Vision and Pattern Recognition (CVPR)},
  pages     = {26296--26306},
  year      = {2024}
}

@misc{liu2024llavanext,
  author       = {Liu, Haotian and Li, Chunyuan and Li, Yuheng and Li, Bo and Zhang, Yuanhan and Shen, Sheng and Lee, Yong Jae},
  title        = {{LLaVA-NeXT}: Improved Reasoning, {OCR}, and World Knowledge},
  howpublished = {LLaVA Blog},
  month        = {January},
  year         = {2024},
  url          = {https://llava-vl.github.io/blog/2024-01-30-llava-next/}
}

@article{wang2024qwen2vl,
  author  = {Wang, Peng and Bai, Shuai and Tan, Sinan and Wang, Shijie and Fan, Zhihao and Bai, Jinze and Chen, Keqin and Liu, Xuejing and Wang, Jialin and Ge, Wenbin and others},
  title   = {{Qwen2-VL}: Enhancing Vision-Language Model's Perception of the World at Any Resolution},
  journal = {arXiv preprint arXiv:2409.12191},
  year    = {2024}
}

@article{qwen2.5vl,
  author  = {Bai, Shuai and Chen, Keqin and Liu, Xuejing and Wang, Jialin and Ge, Wenbin and Song, Sibo and Dang, Kai and Wang, Peng and Wang, Shijie and Tang, Jun and others},
  title   = {{Qwen2.5-VL} Technical Report},
  journal = {arXiv preprint arXiv:2502.13923},
  year    = {2025}
}

@misc{openai2024gpt4o,
  title={Hello {GPT-4o}},
  author={{OpenAI}},
  year={2024},
  url={https://openai.com/index/hello-gpt-4o/}
}

@inproceedings{girshick2015fast,
  author    = {Girshick, Ross},
  title     = {Fast {R-CNN}},
  booktitle = {Proceedings of the IEEE International Conference on Computer Vision (ICCV)},
  year      = {2015},
  pages     = {1440--1448}
}

@article{zhou2014learning,
  author  = {Zhou, B. and Lapedriza, A. and Xiao, J. and Torralba, A. and Oliva, A.},
  title   = {Learning Deep Features for Scene Recognition Using Places Database},
  journal = {Advances in Neural Information Processing Systems},
  volume  = {27},
  year    = {2014}
}

@inproceedings{guo2017calibration,
  author    = {Guo, Chuan and Pleiss, Geoff and Sun, Yu and Weinberger, Kilian Q.},
  title     = {On Calibration of Modern Neural Networks},
  booktitle = {Proceedings of the 34th International Conference on Machine Learning (ICML)},
  series    = {Proceedings of Machine Learning Research},
  volume    = {70},
  pages     = {1321--1330},
  year      = {2017},
  publisher = {PMLR}
}

@inproceedings{nixon2019measuring,
  author    = {Nixon, Jeremy and Dusenberry, Michael W. and Zhang, Linchuan and Jerfel, Ghassen and Tran, Dustin},
  title     = {Measuring Calibration in Deep Learning},
  booktitle = {Proceedings of the IEEE/CVF Conference on Computer Vision and Pattern Recognition Workshops (CVPRW)},
  pages     = {38--41},
  year      = {2019}
}

@inproceedings{machajdik2010affective,
  author    = {Machajdik, Jana and Hanbury, Allan},
  title     = {Affective Image Classification Using Features Inspired by Psychology and Art Theory},
  booktitle = {Proceedings of the 18th ACM International Conference on Multimedia (ACM MM)},
  pages     = {83--92},
  year      = {2010},
  doi       = {10.1145/1873951.1873965}
}

\end{document}